\documentclass[a4paper,11pt]{article}

\usepackage[english]{babel}
\usepackage[T1]{fontenc}
\usepackage[utf8]{inputenc}
\usepackage{setspace}

\usepackage{booktabs}
\usepackage{multirow}	% table with text in multiple rows
\usepackage{rotating}	% rotate text in table

\usepackage[cmex10]{amsmath}
\usepackage{algorithmic}
\usepackage{array}
\usepackage{graphicx}
\usepackage{lmodern}
\usepackage{import}
\usepackage{psfrag}
\usepackage{geometry}                                    
\usepackage[tight,footnotesize]{subfigure}
\usepackage{url}
\usepackage{listings} %Quelltext-Umgebung
\usepackage{color}
\usepackage{units}	% Einheiten mit korrektem Abstand
\usepackage{upgreek}
\usepackage{setspace}
\usepackage{float}
\usepackage{todonotes}
\usepackage{natbib}

\begin{document}

\begin{titlepage}

	\center % Center everything on the page
	
	%----------------------------------------------------------------------------------------
	%   HEADING SECTIONS
	%----------------------------------------------------------------------------------------
	
	%----------------------------------------------------------------------------------------
	%   TITLE SECTION
	%----------------------------------------------------------------------------------------
	\onehalfspacing
	
	%\HRule \\
	{ \LARGE \bfseries  Coherent False Seizure Prediction in Epilepsy, Coincidence or Providence?}\\[0.5cm]

	%----------------------------------------------------------------------------------------
	%   AUTHOR SECTION
	%----------------------------------------------------------------------------------------
	\newcommand\Mark[1]{\textsuperscript#1}
	
	Jens Müller\Mark{1},    
	Hongliu Yang\Mark{1}, Matthias Eberlein\Mark{1}, Georg Leonhardt\Mark{2}, Ortrud Uckermann\Mark{2}, Levin Kuhlmann\Mark{3}, and Ronald Tetzlaff\Mark{1} 	\\[0.5cm]
	
	\raggedright
	
	\Mark{1} TU Dresden, Faculty of Electrical and Computer Engineering, Institute of Circuits and Systems, 01062 Dresden, Germany\\
	\Mark{2} TU Dresden, Neurosurgery of University Hospital Carl Gustav Carus, Fetscherstr. 74, 01307 Dresden, Germany\\
	\Mark{3} Department of Medicine, St Vincent’s Hospital Melbourne, Fitzroy VIC 3065, Australia \\[0.5cm] 
	
\singlespacing
\section*{Abstract}
\paragraph*{Objective}

Seizure forecasting using machine learning is possible, but the performance is far from ideal, as indicated by many false predictions and low specificity. Here, we examine false and missing alarms of two algorithms on long-term datasets to show that the limitations are less related to classifiers or features, but rather to intrinsic changes in the data. 

\paragraph*{Methods}
We evaluated two algorithms on three datasets by computing the correlation of false predictions and estimating the information transfer between both classification methods.

\paragraph*{Results}

For 9 out of 12 individuals both methods showed a performance better than chance. For all individuals we observed a positive correlation in predictions. For individuals with strong correlation in false predictions we were able to boost the performance of one method by excluding test samples based on the results of the second method.

 \paragraph*{Conclusions}

Substantially different algorithms exhibit a highly consistent performance and a strong coherency in false and missing alarms. Hence, changing the underlying hypothesis of a preictal state of fixed time length prior to each seizure to a proictal state is  more helpful  than further optimizing classifiers. 

 \paragraph*{Significance}
The outcome is significant for the evaluation of seizure prediction algorithms on continuous data.
	
\end{titlepage}

\newpage

\section{Introduction}

The prediction of epileptic seizures has become a more and more realistic scenario, promoted by new capabilities and insights into the field of data-driven signal processing in the last decades. For over 30 years, researchers have been trying to identify precursors of seizures and to understand the underlying mechanisms of ictogenesis in the epileptic brain \citep{Kuhlmann2018}. After being too optimistic in the late 1990s on the predictive performance of many approaches that had been derived from short-term intracranial electroencephalography (iEEG) data, the focus moved on to an assessment of new algorithms of continuous multi-day recordings obtained from pre-surgical evaluation \citep{Mormann2007}.
 
In a prospective study with an implanted advisory system -- the first and only in-man clinical trial so far -- it has been shown that seizure prediction in humans is generally possible \citep{Cook2013}. Subsequently, with the availability of continuous long-term recordings that are available to scientists over collaborative platforms \citep{Brinkmann2016, Howbert2014, Wagenaar2015} significant progress was made in terms of sensitivity and specificity of seizure prediction. 

In two online competitions the contestants focused on the development and improvement of refined algorithms to solve a binary classification problem, i.e. to classify data clips as either \textit{preictal} (i.e. a seizure is imminent) or \textit{interictal} \citep{Brinkmann2016, Kuhlmann2018a}. To solve this classification problem, scientists used state-of-the-art machine learning and deep learning methods like random forest trees, support-vector machines (SVM) \citep{Direito2017}, k-nearest neighbours \citep{Ghaderyan2014}, convolutional neural networks \citep{Eberlein2018, Nejedly2019}, and recurrent networks with long-short term memory (LSTM) \citep{Ma2018, Tsiouris2018} among others.

Despite the fact that in many cases the algorithms perform better than a random predictor, they are mostly considered unsuitable for actual clinical use. This is due to a rather low specificity which is characterised by a high number of false positive alarms or by an inappropriate long time in warning \citep{Snyder2008}. Even though the aim is to achieve a high level of sensitivity to detect as many potential dangerous situations as possible, false alarms should be avoided in order to prevent additional psychological stress for the patients by incorrect warnings or unnecessarily long warning times. 
 
However, it is not clear whether an apparently false positive decision  is always actually a \textit{false prediction} or whether it might uncover an epoch of high seizure likelihood that finally did not end up in ictal event - i.e. is a subclinical event. In this contribution we evaluate false alarms in seizure prediction by comparing the outputs of two substantially differing classifiers. Our hypothesis is that false alarms in more than one method imply some intrinsic changes in the iEEG data that cannot be detected by the data-driven methods used and are thus not necessarily the result of a poor classification algorithm. 
% Den letzten Satz verstehe ich nicht.
% @Georg: Eigentlich können wir die Hypothese schon belegen, denn wir zeigen, dass die Fehlalarme offensichtlich nicht auf den Klassifikator zurückzuführen sind. 

%#######################

\section{Materials and Methods}
\subsection{Datasets}
In this study we used three different datasets of long-term intracranial electroencephalography (iEEG). Two of them comprise iEEG segments preselected for recent public seizure challenges, which have become benchmark datasets of seizure prediction studies \citep{Brinkmann2016, Kuhlmann2018a}. In addition, a new dataset (\textit{Dataset 1}) with continuous unselected iEEG recordings was invoked to show that our finding was not biased by the selection procedure of the former.

\paragraph{Dataset 1}
Intracranial EEG from five patients (\textit{Patient A-E}) with pharmacorefractory epilepsy were recorded during presurgical diagnostics using subdural stripes and grids, and depth electrodes at the Department of Neurosurgery of University Hospital Carl Gustav Carus in Dresden, Germany \citep{Eberlein2019}. Beginning and end of each clinical seizure and artefacts were annotated  by an clinical neurologist (G.~L.) specialised in epileptology. For data analysis, between 58 and 107 channels at a sampling rate of \unit[500]{Hz} or \unit[1]{kHz} were available. Patients' details are given in Table~\ref{tab:dresden_data}. 

\begin{table}[ht]
	\centering
	\caption{Characteristics of Dataset 1.
	}
	%	 and \ref{sec:segm_and_label}
	\begin{tabular}{m{20mm}m{10mm}m{10mm}m{20mm}m{15mm}m{15mm}m{22mm}}
		\hline
		\hline
		individual 	& age 	& sex 	& sampling rate	& channels & seizures & recorded duration  \\\hline 
		Patient~A	& 36	& m	& 1000 Hz		& 77 	& 13 	& 161 h \\
		Patient~B	& 38		& m		& 1000 Hz		& 58 	& 3  	& 230 h \\
		Patient~C	& 62		& m		& 500 Hz		& 107   & 7  	& 257 h \\
		Patient~D	& 28		& m		& 500 Hz		& 80    & 27 	& 204 h \\
		Patient~E	& 49		& m		& 500 Hz		& 62  	& 3  	& 139 h \\
		\hline
	\end{tabular}
	\label{tab:dresden_data}
\end{table}

\paragraph{Dataset 2}

Data from the NeuroVista seizure advisory system implant \citep{Coles2013, Davis2011} that had been recorded from human patients and dogs with naturally occurring epilepsy were used. In the context of the \textit{American Epilepsy Society Seizure Prediction Challenge} that had been conducted on the platform kaggle.com, data from five canines and two humans was made available to the contestants  \citep{Brinkmann2016}. From these datasets, we excluded the human patients and \textit{Dog 5} since their data acquisition differs from the other data (\textit{Dog 1} - \textit{Dog 4}), which had been recorded with 16 channels and a sampling rate of \unit[400]{Hz}.

\paragraph{Dataset 3}

The iEEG data was made available to the contestants of the
\textit{Melbourne-University AES-MathWorks-NIH Seizure Prediction Challenge} \citep{Kuhlmann2018a} and includes recordings of three human patients that had taken part in a study with the NeuroVista system \citep{Cook2013}. The iEEG data comprises recordings for about six months occurring after the first month of implantation and was also recorded from 16 electrodes and sampled at \unit[400]{Hz}. 

%\textcolor{red}{The total recording time of each patient exceeds one year, the dataset contains data randomly drawn from a period of about half a year. Here, the placement of the electrode arrays was targeted to the presumed seizure focus, so no detailed information is provided in \citep{Cook2013, Kuhlmann2018} about the relative positioning of the electrode arrays. Again, only preictal data preceding leading seizures was considered following the convention of dataset 1, as described above. The data was also provided as 10-min clips, similar to dataset 1. Interictal data was randomly selected from periods with a minimum gap of \unit[3]{h} before and \unit[4]{h} after any seizure.}

\subsection{Signal Processing}

\subsubsection{General Considerations}
In this contribution, we compare two approaches for seizure prediction that both show a good performance when applied to long-term iEEG datasets but follow fundamentally different strategies. The first method uses a list of "hand-crafted" univariate and bivariate features that have proved suitable for EEG characterisation \citep{Tetzlaff2012, Senger2016}. Feature vectors were classified by a multilayer perceptron (MLP) and the best feature combination is chosen to optimise the performance of the algorithm. The second method applies a deep neural network to the iEEG raw data. By means of subsequent convolution and pooling layers the feature extraction and classification is taken over completely by a deep learning process.

\paragraph*{Data Handling}
All recordings were divided into a \textit{training set} to fit our models and a \textit{test set} for the evaluation on out-of-sample data. For each individual, training and test sets were separated in time. In case of the five patients of \textit{Dataset 1} the first half of the data was assigned as training data and the second half as test data \citep{Eberlein2019}. Details about the structure of \textit{Dataset 2} and \textit{Dataset 3} are given in \citep{Korshunova2018} and \citep{Kuhlmann2018a}.

\paragraph*{Segmentation of Dataset 1}
The segmentation of recordings from \textit{Dataset 1} was done in general accordance with the procedure of the competitions on kaggle.com as outlined in \citep{Brinkmann2016} and \citep{Kuhlmann2018a}. All data was divided into contiguous, non-overlapping 10-min clips. The data from \unit[66]{min} to \unit[5]{min} before onset of a seizure was assigned as \textit{preictal}. The time period of 60 min following a seizure onset was excluded from the analysis to avoid derogation of the data by ictal and postictal behaviour. Any seizure that might have occurred during that time period would have not been included in the analysis.  Data that was recorded at least 4~h from any seizures (i.e. from 240 min after a seizure till 240 min before the next seizure) was assigned as \textit{interictal}.

To avoid data contamination by events exceeding the recorded duration, \unit[4]{h} of data at the beginning and at the end of each recording was discarded. EEG channels showing artefacts (identified by visual inspection by the neurologist) were excluded from this study. Table~\ref{tab:datasets_2} provides an overview to the numbers of  interictal and preictal clips of all datasets.

\paragraph*{Preprocessing}
All data was subsampled to \unit[200]{Hz} in order to reduce computational cost. Before being fed into the CNN,  z-score normalisation was applied on each channel individually all 10-min clips. Subsequently, the 10~min sequences were divided in segments of \unit[15]{s}. In this study, an adaptive-training approach with retraining after a fixed period or seizure event is not considered.

\begin{table}[ht]
	\renewcommand{\arraystretch}{1.3}
	\caption{Number of interictal and preictal 10-min clips for all three datasets.}
	\label{tab:datasets_2}
	\begin{center}
		\vspace{-0.3cm}
		\begin{tabular}{ >{\centering\arraybackslash} p{1.7cm} >{\centering\arraybackslash} p{1.5cm} >{\centering\arraybackslash} p{1.5cm} >{\centering\arraybackslash} p{1.5cm} >{\centering\arraybackslash} p{1.5cm} }
			\hline
			\hline
			&	 \multicolumn{2}{c}{training clips}  & \multicolumn{2}{c}{testing clips}  \\
			&	 interictal & preictal & interictal & preictal  	\\		
			\hline
			Patient A & 203 & 36 & 351 & 25 \\ 
			Patient B & 760 & 12 & 424 & 6 \\ 
			Patient C & 416 & 28 & 828 & 12 \\ 
			Patient D & 236 & 74 & 206 & 27 \\ 
			Patient E & 272 & 7 & 408 & 6 \\ 
			\hline
			Dog 1 & 480 & 24 & 478 & 24 \\ 
			Dog 2 & 500 & 42 & 910 & 90 \\ 
			Dog 3 & 1440 & 72 & 865 & 42 \\ 
			Dog 4 & 804 & 97 & 933 & 57 \\ 
			\hline
			Patient 1 & 570 & 256 &  156 & 60  \\ 
			Patient 2 & 1836 & 222 &  942 & 60  \\ 
			Patient 3 & 1908 & 255 &  630 & 60  \\ 
			\hline
		\end{tabular}
	\end{center}
	\vspace*{-0.25cm}	
\end{table}

\subsubsection{Feature-based Classification (Method 1)}
%%%%%%%%%%%%%%%%%%%%%%%%%%%%%%%%%%%%%%%%%%%%%%%%%%%%%%%%%%%%%
\paragraph*{Features}
We considered both univariate and bivariate features to account for epileptiform anomalies in single-channel iEEG measurements and among their correlations. A major part of our core feature set comes from top ranked algorithms of the American Epilepsy Society Seizure Prediction Challenge on kaggle.com \citep{Brinkmann2016}. This includes band power spectrum and statistical moments of single channel signals as well as cross-channel (linear) correlations in time and frequency domains. Other methods were adapted from previous studies to complement and strengthen the set further \citep{Mormann2007, Tetzlaff2012}. For univariate features we added autoregressive (AR) model coefficients and signal prediction errors, which capture sequential information of iEEG signals \citep{Tetzlaff2012}. 
%In total, 23 univariate features are considered.

For bivariate features we used nonlinear interdependence\citep{Andrzejak2003} and mean phase coherence\citep{Mormann2003} to characterise nonlinear cross-channel coherence. In view of the complex nature of brain dynamics, nonlinear measures are expected to be more suitable for extracting information from iEEG signals. We considered three variants of symmetric bivariate features: 1) the feature matrix itself, 2) eigenvalues and eigenvectors of the matrix as well as the maximum of rows/columns, and 3) a combination of all features of the two previous variants. 
%In total, this amounts to 22 bivariate features. 

%\begin{table}[t]				% TABLE I
%	\renewcommand{\arraystretch}{1.3}
%\caption{
%List of considered univariate and bivariate core features.
%} 			
%\thispagestyle{plain}
%\centering{						
%\begin{tabular}{l  | r } 				
%\hline                     			
%\hline                     			
%univariate features  & bivariate features\\ [0.5ex] 
%\hline                  				
%Coefficients of AR model	& Granger causality\\
%Mean pred error of AR model  & Nonlinear interdependence\\
%Mean and SD of pred error & Mean phase coherence\\
%Band power spectrum  & Correlation coeff. in time\\
%Skewness \& kurtosis & Linear coherence\\
% & Corr. coeff. of signal amplitude/phase\\[1ex]
%\hline 							
%\end{tabular}
%}\\
%\label{table:features} 					
%\end{table}

\paragraph*{Classification}
The classification was executed on input vectors of each univariate and bivariate feature by means of a multilayer perceptron (MLP) with three hidden layers comprising 16, 8, and 4 neurons, respectively. The combination of uni- and bivariate characteristics is intended to provide both properties of single channels and of their correlations. Each network layer is followed by a batch normalization. The rectified linear unit (ReLU) activation function is used for hidden layers and  the sigmoid function for the output layer. The network was trained with stochastic gradient descent (SGD) using backpropagation over 500 epochs with a learning rate of $10^{-4}$. It was found that dropout does not improve the performance of such a small network. We considered the mean of an ensemble of 100 networks with different initial weights in order to obtain a statistical significance of the outputs. 

%\begin{figure}[t]
%	\centering
%	\includegraphics[width=0.7\columnwidth]{patient2_roc_valid.eps}
%	\caption{ROC AUC values of all feature combinations of \textit{Patient 2} of \textit{Dataset 3} estimated on a validation set. In indices on the axes represent a specific univariate/multivariate feature of Table~\ref{table:features}.}
%	\label{fig:kaggle16-roc-mean-pat2}
%\end{figure}

To find the optimal feature combination the area under the receiver operating characteristic curve (ROC AUC value) of all possible combinations was estimated on a respective validation set and the feature combination with the highest validation score was selected as the optimal one to get the test scores and predictions. In case of the patients from \textit{Dataset 1},  the validation set equals the test set due to the relatively short recording time for each patient. For \textit{Dataset 2} and \textit{Dataset 3} the available public test sets were chosen for validation. We are fully aware of the limits of significance of our methodology and that the performance does not reflect a purely prospective approach, where the optimization of the model's hyperparameters should be done on the training data \citep{Eberlein2018}. However, we accept this in the context of this study since the focus is directed to the false alarms and less on the prediction performance itself. 

%An example plot of the validation score is shown in Fig.~\ref{fig:kaggle16-roc-mean-pat2} for \textit{Patient 2} of \textit{Dataset 3}. 
%%%%%%%%%%%%%%%%%%%%%%%%%%%%%%%%%%%%%%%%%%%%%%%%%%%%%%%%%%%%%

\subsubsection{Deep-Learning Classification (Method 2)}

In comparison to the "hand-crafted" feature extraction we applied a convolutional neural network to the multi-channel iEEG data, as originally proposed as \textit{topology 1 (nv1x16)} in \citep{Eberlein2018}. An appropriate low-dimensional representation of the signal was derived from the raw data in recurring consecutive layers of convolution, nonlinear activation, and pooling operations. 

\begin{figure}[H]
	\centering
	\includegraphics[width=\columnwidth]{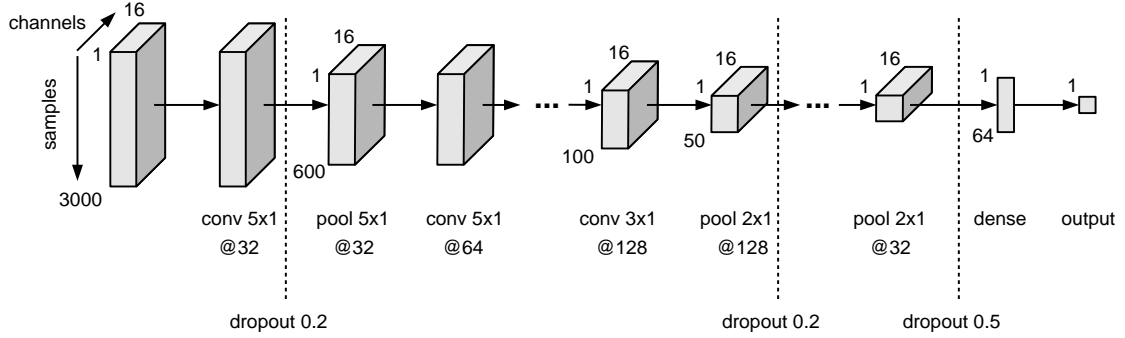}
	\caption{Schematic of the convolutional neural network (CNN) topology to process multi-channel iEEG raw data, exemplarily shown for 16-channel input data with 3,000 samples.}
	\label{fig:cnn-topology}
\end{figure}

A schematic of the topology is depicted in Fig.~\ref{fig:cnn-topology}. The input data (a $k$-channel array of 3,000 samples) is processed along the time axis by convolution and pooling operations with kernel sizes in the range of 2 to 5. The number of feature maps changes from 32 in the first layers to 128 and to 32 again in the last layers. ReLU was chosen as nonlinear activation in the convolutional and dense layers and the sigmoid function was used in the output layer. Additionally, dropout layers ($p=0.2$ and $p=0.5$) as well as L1 and L2 regularization were applied. Finally, the classification was done in a fully connected layer of 64 neurons.

In contrast to method 1, the network's hyperparameters were optimised by using training and test data of \textit{Dog 2} to avoid an overfitting of the models. For all remaining individuals, the derived network topology is applied without using further validation sets.
To improve the statistical significance 20 models were trained for each individual.

\subsubsection{Correlation of prediction errors}
Various metrics have been used for assessing the prediction performance, for instance accuracy, sensitivity, etc. Here, we are interested in changes of predictions over time and especially in the consistency of changes between two different methods. Subsequently, the term \textit{prediction} is considered as the respective network output and represents the probability of an iEEG clip to be preictal. According to our hypothesis, a high correlation of predictions of fundamentally different classifiers indicates an intrinsic change in the data.

A standard measure used to characterise the coherence between two data series is the pearson correlation coefficient. 
For two predictions $\vec{p}_{i}$ and $\vec{p}_{j}$ it reads
\begin{equation}
\label{eq:c}
c = \frac{\langle (\vec{p}_i- \langle \vec{p}_i\rangle) (\vec{p}_j- \langle \vec{p}_j\rangle) \rangle }{ \sigma(\vec{p}_i)\sigma(\vec{p}_j)},
\end{equation}
where $\langle \cdot \rangle$ and $\sigma(\cdot)$ denote the mean and standard deviation of a prediction for all samples. Since any good prediction should be close to the ground truth the correlation $c$ is strongly biased, i.e. its value is almost trivially high for two good predictions.

For a metric with emphasis on the coherence between "false" predictions we used the \textit{weighted} correlation coefficient
\begin{equation}
\label{eq:c_w}
c_w = \frac{\langle (\vec{e}_i- \langle \vec{e}_i\rangle_w) (\vec{e}_j- \langle \vec{e}_j\rangle_w) \rangle_w }{ \sigma_w(\vec{e}_i)\sigma_w(\vec{e}_j)},
\end{equation}
where $\vec{e}_i=\left| \vec{p}_{i} -\vec{L}\right|$ denotes the prediction error and $\vec{L}$ is the label. The weighted mean and standard deviation are given by
\begin{equation}
\langle z \rangle_w = \frac{ \sum _k w_k z_k } {\sum _k w_k },
\end{equation}
\begin{equation}
\sigma_w (z) = \sqrt {\langle z^2 \rangle_w - \langle z \rangle^2_w}.
\end{equation}
The weight factor $w= \max{(\vec{e}_i, \vec{e}_j)}$ was assigned to emphasize the effect of coherent false predictions.

\subsubsection{Information transfer}
As another demonstration of the observed coherent false predictions, we tested how knowledge of false prediction obtained for method A can be used to artificially boost the performance of method B and vice versa. To be precise, with predictions from method A we eliminate all clips with the prediction error $e$ larger than a threshold value $e_\mathrm{th}$. Then the ROC AUC value of method B is evaluated on the reduced dataset. If there is a strong correlation of the prediction errors, we expect that samples which are difficult to be classified for method A are likely to be classified falsely by method B as well. 

\section{Results}

\subsection{Classification performance}
For a non-subjective characterization of the performance of our classifiers we used the statistical metric of the ROC AUC value. In order to obtain classifications of the original \unit[10]{min} clips, the predictions of the corresponding \unit[15]{s} segments have been averaged.  Due to the missing timestamps of \textit{Dataset 2} and \textit{Dataset 3}, it was not possible to calculate other metrics on specificity like \textit{time in warning} \citep{Mormann2007}.  As shown  in Tab.~\ref{table:auc-corr},  both classifiers performed better than a chance perdictor in 9 out of 12 individuals from all three datasets. The performance is comparable to the state-of-the-art leading algorithms for the long-term recordings of \textit{Dataset 2} \citep{Brinkmann2016} and \textit{Dataset 3} \citep{Kuhlmann2018a} 

%\textit{das ist zu allgemein, da müssen genauere Daten her --> Vergleichswerte sind in Tabelle eingetragen}. 

%Limited by the amount of data available for training, our classifiers did not achieve the same level of performance for Patients B and D of Dataset 1 as for other individuals. They therefore deserve a separate discussion in the following sections.

%\begin{table}[ht]				% TABLE IV
%	\renewcommand{\arraystretch}{1.3}
%\caption{
%Seizure prediction performance of the two methods characterized by ROC AUC values. 
%} 			
%\thispagestyle{plain}
%\centering{						
%\begin{tabular}{c c c c c c c c c c  c c c} 				
%\hline                     			
%\hline                     			
%\multirow{2}{*}{individual} & \multicolumn{5}{c}{Dataset 1} & \multicolumn{4}{c}{Dataset 2} & \multicolumn{3}{c}{Dataset 3} \\
%\cmidrule(lr){2-6} \cmidrule(lr){7-10} \cmidrule(lr){11-13}
%& Pat A & Pat B & Pat C & Pat D & Pat E  &Dog1 & Dog2 & Dog3 & Dog4 & Pat1 & Pat2 & Pat3 \\ [0.5ex] 
%\hline                  				
%Method 1 & 0.79 & 0.46 & 0.91 & 0.65 & 0.96  & 0.92 & 0.85 & 0.82 & 0.90 & 0.78 & 0.87 & 0.82\\
%Method 2 & 0.74 & 0.25 & 0.87 & 0.32 & 0.82  & 0.80 & 0.81 & 0.84 & 0.92 & 0.25 & 0.75 & 0.77
%\\[1ex]
%\hline 							
%\end{tabular}
%}\\
%\label{table:aucs} 					
%\end{table}

\subsection{Coherent false predictions and information transfer}
%\textit{die ersten beiden Sätze verstehe ich nicht: }Much attention has been paid on various statistical metrics when comparing the classification methods although it is known that the detailed prediction time series itself contain even richer information.

\begin{figure}[H]
	\centering
	\includegraphics[width=0.85\columnwidth]{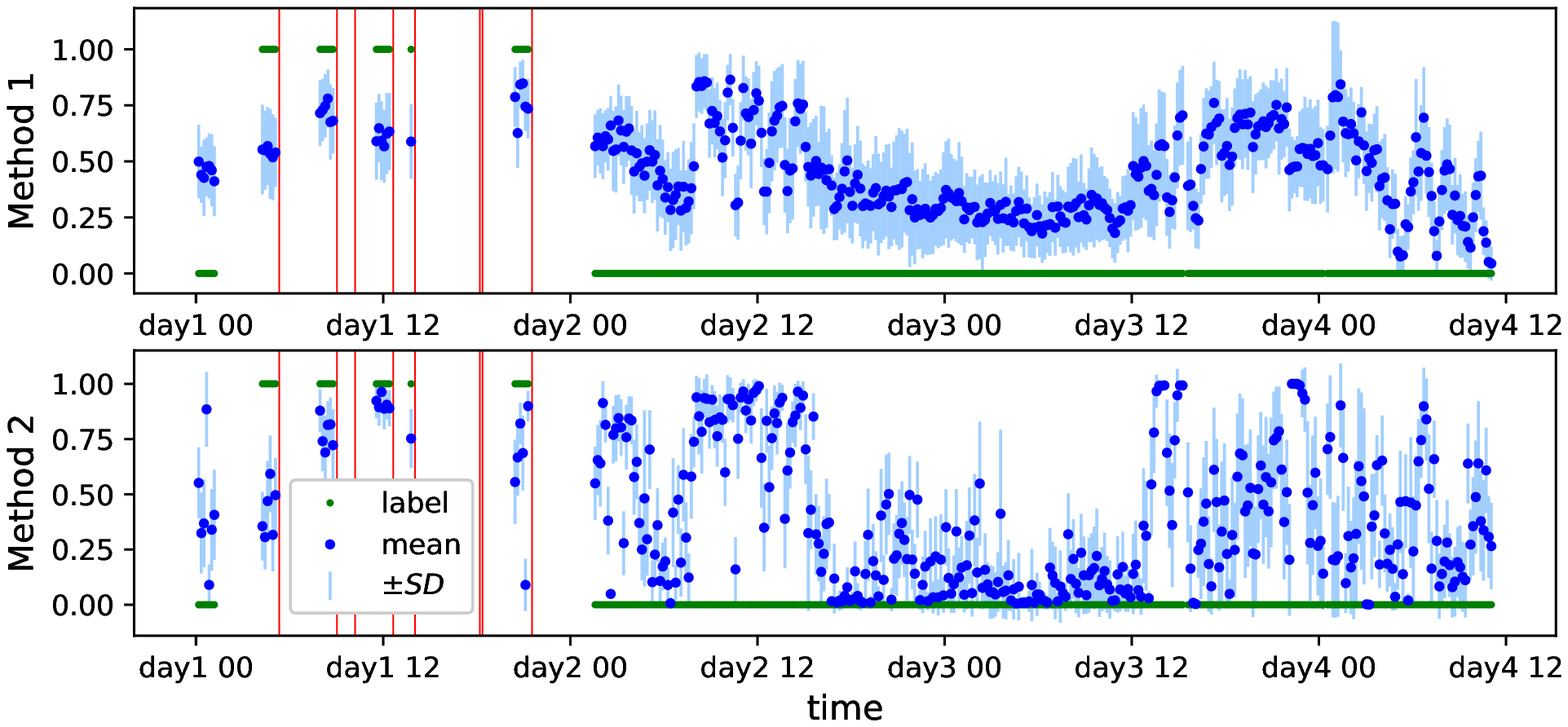}
	\includegraphics[width=0.85\columnwidth]{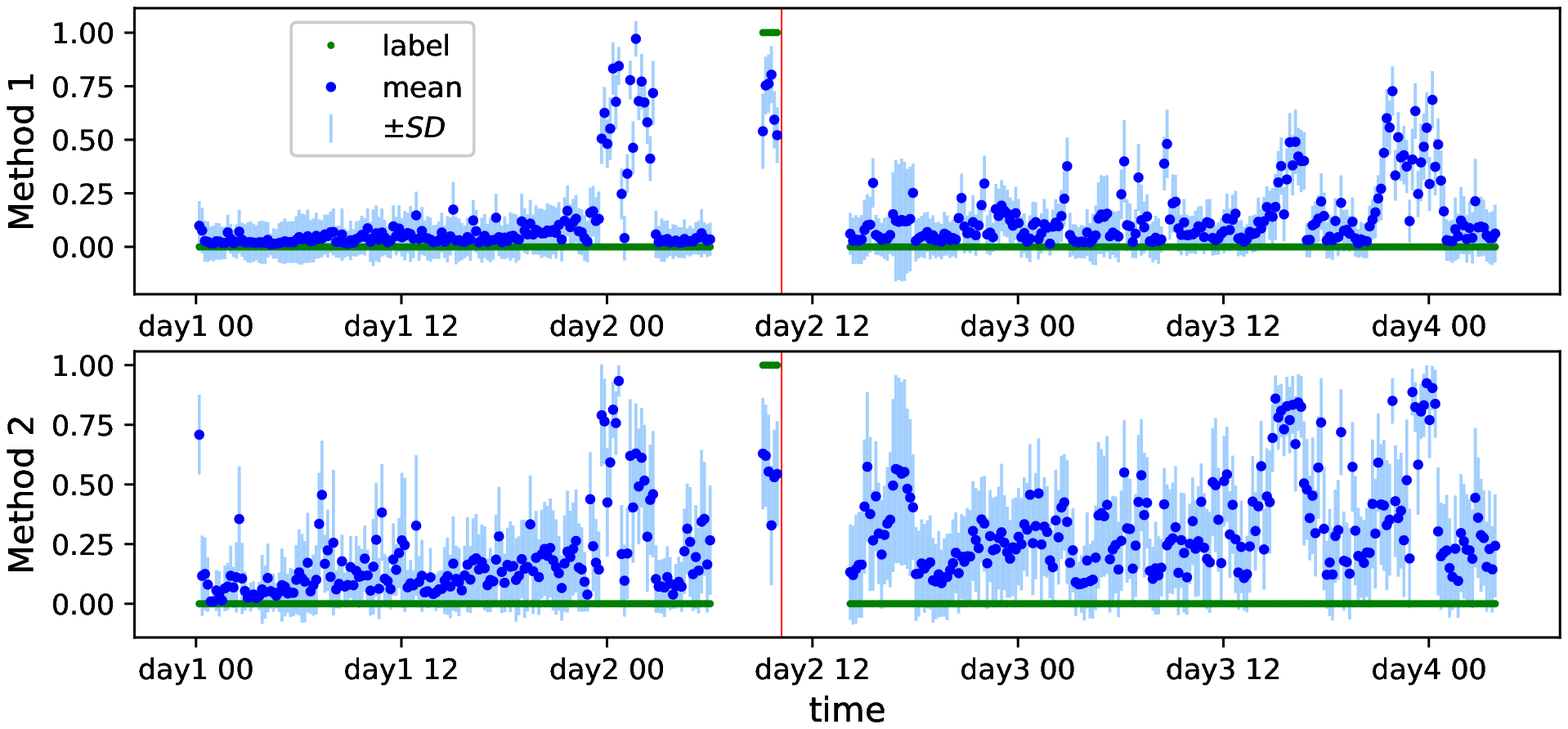}
	\caption{Comparison of predictions in the testing data for\textit{ Patient A} (upper) and \textit{Patient E} (lower). The standard deviation (SD) is computed over 100 models for method 1 and 20 models for method 2. False positive predictions are delivered coherently by both methods for instance for \textit{Patient A} from day 2 8:00 am to 4:00 pm and for \textit{Patient E} from day 2 0:00 am to 3:00 am. Seizure onset is marked as red line.}
	\label{fig:ukdd-fea-cnn-err-pat1}
\end{figure}

Interesting new aspects were found when comparing the prediction time series directly, as shown exemplarily for \textit{Patient A} and \textit{Patient E} in Fig.~\ref{fig:ukdd-fea-cnn-err-pat1}. Both classifiers delivered a good performance when applied to the recordings of \textit{Patient A} as characterised by AUC values of 0.79 for method 1 and 0.74 for method 2. Hence, it is not surprising to see that both methods made correct predictions consistently for many segments. Noticeably, they delivered false predictions also in a highly coherent manner, as indicated for instance in the cluster of false positive states from day 2 8:00 am to 4:00 pm. Moreover, the way how the prediction values change with time is also very coherent. The same phenomenon  can be seen  at the predictions over time of \textit{Patient E} in the lower part of Fig.~\ref{fig:ukdd-fea-cnn-err-pat1}, where a strong coherence of false predictions is observed for example on day 2 from 0:00 am to 3:00 am.

% Not into discussion, since it's a general explanation
The correlation coefficient $c$ and weighted correlation coefficient $c_w$  were used to measure the observed coherency between false predictions. Here, negative values represent anti-correlation, meaning a state falsely labelled by classifier A will be correctly labelled by classifier B and vice versa. The positive values of $c$ and $c_w$ (see Tab.~\ref{table:auc-corr}) indicate that the two classifiers are mistaken about the same time periods by giving coherent false predictions, with a higher positive value representing a stronger coherency. 

Note that for \textit{Patient C} of \textit{Dataset 1} the coherent false predictions have a clear one-day rhythm (see Fig.S1-S4 in the supplementary material), which evidences a direct cause from the circadian cycle. However, no regular cycles are visible for other patients of \textit{Dataset 1} with a significantly high coherence measure $c_w$. Hence, circadian cycle and other periodicity can not be the only cause of the observed coherency in false predictions.

\begin{table}[H]				% TABLE IV
	\renewcommand{\arraystretch}{1.3}
	\caption{
		Seizure prediction performance of the two methods characterised by receiver operating characteristic (ROC) area under curve (AUC) values and correlation of the predictions $c$ and the weighted correlation of prediction errors $c_w$. Both methods were compared against random predictors and p-values of their superior performance were assessed using the Hanley-McNeil method \citep{Hanley1982}. For the correlations, p-values are estimated as the probability of two random predictions with the same ROC AUC having equal or higher values. A total of $M=100$ random predictions were obtained for each method by randomly permuting the original predictions for each label class individually. All possible combinations amount to $N=5050$ correlations for computing the p-values. For $c$ and $c_w$, all p-values are $< 1/N$, except for \textsuperscript{a}: $p = 0.0008$, \textsuperscript{b}: $p=0.008$, \textsuperscript{c}: $p=0.02$} 		
	\thispagestyle{plain}
	\centering{						
		\begin{tabular}{c l | c c | c c | c c | m{3.5cm}}
			\hline                     			
			\hline 	
			&	&	\multicolumn{2}{m{2cm}|}{\centering Method 1} & \multicolumn{2}{m{2cm}|}{\centering Method 2}	 & \multicolumn{2}{m{2.5cm}|}{\centering correlation of predictions} &  winning solution in \citep{Brinkmann2016} and \citep{Kuhlmann2018a} 	\\
			&	& AUC  & $p$ &	AUC & $p$   & $c$ &	$c_w$ & AUC\\
			\hline
			\multirow{5}{*}{\begin{sideways} Dataset 1 \end{sideways}} 
			& Pat. A & 0.83 & < 0.001 & 0.77 & < 0.001 & 0.66 & 0.56 & -\\
			& Pat. B & 0.23 & 1.00 & 0.22 & 1.00 & 0.23 & 0.03\textsuperscript{c} & -\\
			& Pat. C & 0.89 & < 0.001 & 0.88 & < 0.001 & 0.92 & 0.84 & -\\
			& Pat. D & 0.63 & 0.01 & 0.33 & 1.00 & 0.49 & 0.35 & -\\
			& Pat. E & 0.97 & < 0.001 & 0.88 & < 0.001 & 0.70 & 0.64 & -\\
			\hline                    			
			\multirow{4}{*}{\begin{sideways} Dataset 2 \end{sideways}} 	
			& Dog 1	& 0.97 & < 0.001 & 0.82 & < 0.001 & 0.69 & 0.54 & 0.94\\
			& Dog 2	& 0.89 & < 0.001 & 0.80 & < 0.001 & 0.71 & 0.63 & 0.86\\
			& Dog 3	& 0.89 & < 0.001 & 0.83 & < 0.001 & 0.71 & 0.79 & 0.86\\
			& Dog 4	& 0.93 & < 0.001 & 0.90 & < 0.001 & 0.73 & 0.49 & 0.89\\
			\hline                    			
			\multirow{3}{*}{\begin{sideways} Dataset 3 \end{sideways}} 	
			& Pat. 1	& 0.63 & 0.001 & 0.24 & 1.00 & 0.14\textsuperscript{a} & 0.04\textsuperscript{b} & 0.55\\
			& Pat. 2	& 0.82 & < 0.001 & 0.74 & < 0.001 & 0.81 & 0.65 & 0.74\\
			& Pat. 2	& 0.83 & < 0.001 & 0.75 & < 0.001 & 0.37 & 0.20 & 0.87\\
			\hline 							
		\end{tabular}
	}\\
	\label{table:auc-corr} 					
\end{table}

For most individuals of all three datasets the values of $c_w$ are larger than 0.5, indicating the common occurrence of a medium to strong correlation between false predictions of two classifiers.  Exceptions are \textit{Patients B}, \textit{Patient D}, and \textit{Patient 1}, where at least one of our classifiers performs significantly worse than the other.  For all individuals the pearson correlation coefficient $c$ is always larger than the corresponding value of $c_w$, which reflects the biasing effect of the common correct predictions.

%subsection{Information transfer}

\begin{figure}[t]
	\centering
	\subfigure[Dog 3 of dataset 2, $c_w=0.79$]{\label{fig:inf-trans-dog3}\includegraphics[width=0.49\columnwidth]{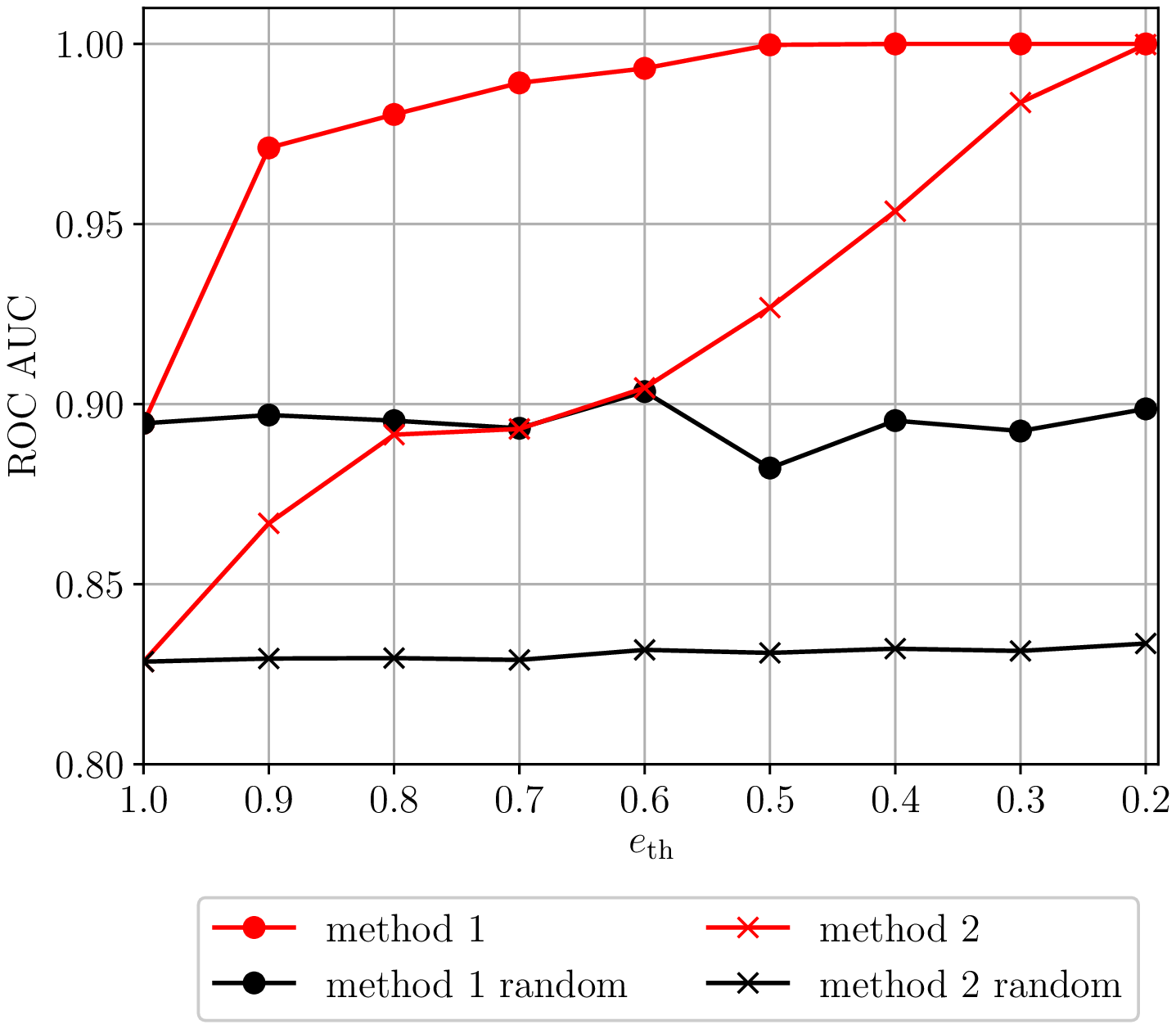}}\hfill
	\subfigure[Patient 1 of dataset 3, $c_w=-0.07$]{\label{fig:inf-trans-pat1}\includegraphics[width=0.49\columnwidth]{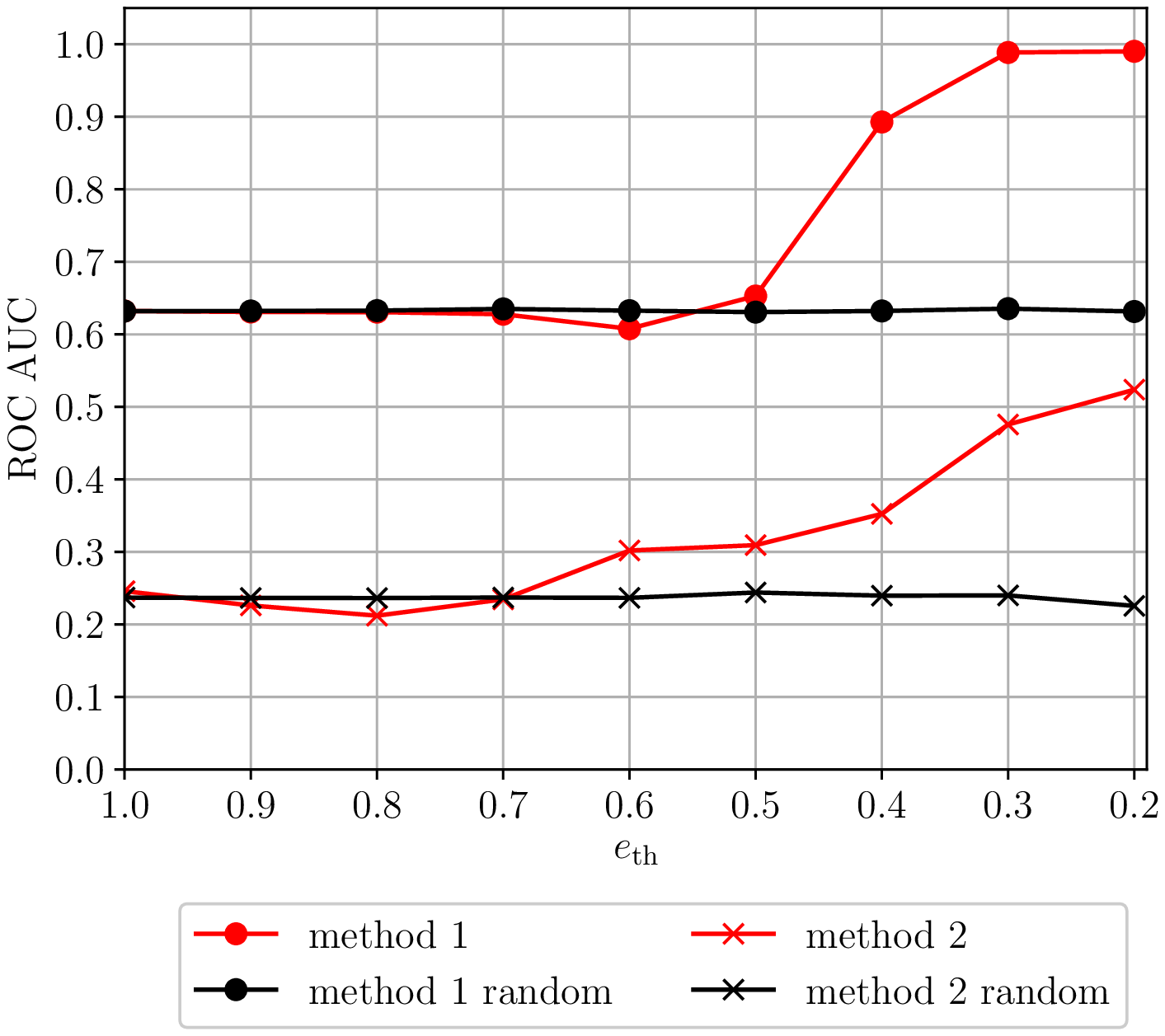}}
	\caption{Coherence of false predictions demonstrated by information transfer between two methods, where $c_w$ is the weighted correlation of prediction errors, $e$ denotes the prediction error, and $e_\mathrm{th}$ denotes the threshold of omission.  Here, "method 1" means that the falsely predicted samples (with $e>e_\mathrm{th}$) of method 1 were eliminated from the test set of method 2, and vice versa. For comparison, "method 1 random" shows the performance for a randomly reduced test set of the same size.}
	\label{fig:inf-transf}
\end{figure}

The change of ROC AUC values depending on the threshold of omission $e_\mathrm{th}$ is shown in Fig.~\ref{fig:inf-trans-dog3} for \textit{Dog 3} and in (b) for \textit{Patient 1}. The ROC AUC value at $e_\mathrm{th}=1$ corresponds to that of the original complete testing set and with decreasing $e_\mathrm{th}$ more and more falsely predicted samples were omitted.

For \textit{Dog 3} in Fig.~\ref{fig:inf-trans-dog3} we clearly observe an increase in the ROC AUC values with decreasing $e_\mathrm{th}$, which is in good correspondence with the strong correlation of false predictions for this individual ($c=0.80$ and $c_w=0.79$, see Tab.~\ref{table:auc-corr}). To exclude the possibility that the increase of ROC AUC values is only due to the decreased amount of testing samples, the performance has been evaluated for a reduced test set with the same number of randomly selected samples being omitted. In this case, the ROC AUC values remains almost constant.

For comparison, we observe no significant increase of the ROC AUC values for \textit{Patient 1} in Fig.~\ref{fig:inf-trans-pat1} as long as $e_\mathrm{th}$  does not fall below a threshold of 0.6. This behaviour can be expected since \textit{Patient 1} shows a rather low correlation of false predictions ($c=0.22$ and $c_w=-0.07$) and hence, it is not likely that the omission of samples falsely predicted by method A will significantly affect the performance of method B. 

\section{Discussion}
%\textit{Georg: I have not yet edited this section completely. I will do that after our discussion.
%Just a few suggestions about structure and content:
%1. paragraph: Summary of what we found especially which mistakes (pos or neg, in the same person or not etc)
%2. par: do we need to discuss the choice of the two methods or has that been done sufficiently in the introduction section ?
%3. par: focus to the mistakes. More specific than on par 1 in which individuals, which direction, more in meth 1 or 2 ... 
%4. par: mistakes in pat of dataset 1: which time of day, were the two pat implanted in the same way and different from the others? 
%5. par Do we have acces to these information in the other pats? Is the number of contacts an issue? Why no mistakes in the dog-data? (long duration of recording, fewer contacts?}

% as a general remark: no colloquial terms such as "reasonable", "clear correlation" etc. either leave them out or use more precise terms. The discusion should be more tight (knapp). If it is structured by subheadings points should not be repeated in the various subheadings.
 
 In our study we were able to show that seizure prediction is possible with a performance better than chance for a majority of the individuals,  substantiated by ROC AUC values above 0.5 for 11 out of 12 individuals for method 1 and for 9 out of 12 individuals for method 2. This result is in line with recent studies \citep{Brinkmann2016} and \citep{Kuhlmann2018a}. 
 
 %\textit{wir wollen doch nicht die allgemeine Stärke der Vorhersage der beiden Methosen vergleichen, sondern Gründe für die false positiven findings finden / diskutieren, oder ??} Daher ist für mich dieser Absatz nicht sinnvoll. --> Es schadet aber auch nicht, dass wir noch einmal kurz auf diesen Punkt hinweisen!

The performance of method 1 is slightly better than method 2 for almost all individuals (except for \textit{Dog 3} and \textit{Dog 4}) as the best feature combination was chosen retrospectively from the validation set. In a prospective approach such an optimisation will not be possible and the performance of this method is therefore expected to be worse , as shown in \citep{Eberlein2019}. Moreover, the relevance of the ROC AUC values of \textit{Dataset 1} is limited since the amount of iEEG recordings is considerably shorter than that of \textit{Dataset 2} and \textit{Dataset 3}, respectively. However, it provides continuous and annotated data which is valuable for the discussion of coherent false predictions given below.

Generally, it seems that the performance for each individual is limited by a "ceiling effect". This is consistent with a recent study that observed that ensembling of top-performing algorithms shows no real improvement \citep{Reuben2019}. These findings imply that classifiers might not be able to perform significantly better on this data and that false predictions are correlated across different algorithms.
 
We assume that the reason for the upper boundary is due to the non-stationarity of the signal that causes temporally varying distributions of the raw data and implies different training and testing distributions for clinically relevant applications, i.e. temporally separated training and test phases. This results in intrinsically limited ability to generalise between those two data sets by means of the data itself. For data driven methods, this deficit could be overcome by significantly more and/or less correlated training data.
 
 \subsection*{Origins of coherent false predictions}
 
 In our analysis on three different long-term datasets, two fundamentally different algorithms show a remarkable coherence in correct and wrong predictions of iEEG sequences, indicated by the weighted correlation coefficient $c_w > 0.5$ in seven out of 12 individuals. For three out of the five remaining individuals (with $c_w < 0.5$) at least one algorithm yields a very weak prediction performance. By looking at the network outputs over time, (as exemplarily given for \textit{Patient A} in Fig.~\ref{fig:ukdd-fea-cnn-err-pat1}) we can observe this correlation as a temporal conformity of the network outputs on the time scale of hours to days. Moreover, our investigation about information transfer reveals an increase in specificity of one method if we eliminate samples from the test set that have been classified falsely by another method (see Fig.~\ref{fig:inf-transf}).
 
%However, the frequent occurrence of coherent wrong predictions raises the question about the origin of the underlying changes in the data. Obviously, the classifiers are sensitive to spatio-temporal alterations in the iEEG but it is still unclear why both models change uniformly in some periods of time.

%\subsubsection*{Model-related}
In this study (as in the majority of similar studies \citep{Brinkmann2016, Kuhlmann2018, Kuhlmann2018a, Mormann2016}), the complex problem of seizure prediction is reduced to a binary classification task of allocating data as either preictal or as interictal. We assume that data-driven algorithms are generally sensitive to patterns that are correlated with these two states of the brain, at least in the training set. However, it is obvious that we observe a variety of different  brain modes, even in interictal periods of the epileptic brain \citep{Kalitzin2011}. Therefore, especially in scenarios with limited and non-stationary data, it is unclear whether all possible states are sufficiently represented in the training data. This leads to two possible interpretations for the causes of frequent occurrence of coherent false predictions.

On the one hand, we expect an overfitting of the models. As already discussed, generalizability is intrinsically limited for the problem at hand. Reasons include the above described long-term non-stationarity or the low number of seizures. Furthermore,  patients might experience different types of seizures with different generating dynamics \citep{Freestone2017, Mormann2005}.   Seizure onsets varying in time and manifesting over different channels were observed for example in \citep{Ung2016}. Potentially seizures of a type that did not occur during the training period are difficult to be predicted by an optimized algorithm. Phenomena like this might explain why the generalization problems not only occur regardless of the algorithm but also generally for the same periods.  This leads to the conclusion, that these problems will persist for data-driven approaches and the present data, regardless of the method applied.

%To avoid such an overfitting an adaption of the model parameters by recurring retraining  might be helpful, as it was reported in \citep{Nejedly2019}.

%\subsubsection{Problem-related}
On the other hand we might not have a valid ground truth for the commonly used working hypothesis of binary classification. The labeling and classification of the EEG segments is based on the fact that a seizure occurred subsequently or not. Given the fact that changes in the EEG-signal that harbour the potential to lead to a seizure but are not yet strong enough to stride over the threshold will not be classified correctly as \textit{preictal} despite its ictogenic potential. In such a scenario the term \textit{proictal} would be more appropriate as it is devoid of the actual occurrence of an apparent seizure.

 %xxx Sorry,  den folgenden Absatz habe ich nicht verstanden. Möchtest Du ihn behalten? An alternative reason for these results could be that the underlying ground truth is incorrect. In this scenario, the coherent wrong predictions are indeed related to relevant clinical manifestations, at least according to the assumptions and the data the algorithm was trained on.  This means that they could be caused by specific choices of parameters during the design of the proposed method. For example, if the assumed seizure prediction horizon is too long and includes dynamics that are actually not associated with seizures, it makes the algorithms susceptible to false positives. However, a systematic examination with prediction horizons of different lengths is still needed.

% xxx Further, the ground truth could change between the training and the test periods due to the brain's self-regulatory mechanisms  \citep{Freestone2017} enabling proictal states to lead to seizures or not .  
 
 Finally, false \textit{negatives} are likely to show up in a set-up with an assumed preictal period of a fixed duration. In accordance with the definitions used in \citep{Brinkmann2016} and \citep{Kuhlmann2018a} we assumed a preictal state in the time between 65 minutes and 5 minutes prior to each seizure. It is however questionable whether such a fixed seizure prediction horizon (SPH) is valid for all individuals or even for all seizures of an individual at all \citep{Snyder2008}. Several studies have already shown that the best performance of prediction algorithms is achieved for a patient-individual SPH in the range of 10~min to 60~min \citep{Gadhoumi2016, Senger2016, Zheng2014} or in even shorter prediction horizons of less than 10 min \citep{Kuhlmann2010}. In our opinion, more flexibility should be provided at this point when considering databases with long-term recordings.

% \todo[inline, color=blue]{the classifier learned to distinguish other changes for instance sleep and wake instead of ii and pi?}

 \subsection*{Steps to improve seizure prediction}

 Regardless of the interpretation of the causes of the presented results, it is likely that we need a  new working hypothesis, since the binary classification of interictal and preictal segments is limited. The assumption of a \textit{preictal} state is based on the assumption of a deterministic transition from the interictal state to a seizure. Physiological or pathophysiological processes that are related to ictogenesis but not inevitably followed by a seizure are not considered in this hypothesis. Recently, a more probabilistic approach is often considered by assuming the existence of a \textit{proictal} state that is characterised by an increased probability of a seizure onset \citep{Kalitzin2011, Meisel2015}. Studies on forecasting epileptic seizures that identify periods of increased risk of seizures based on the analysis of circadian and multi-day cycles already show promising results,  \citep{Karoly2017, Proix2019, Stirling2020}.
 
However, the retrospective determination of increased seizure risk is unfeasible for currently available data. This is the case especially for periods of high seizure likelihood that are not developing into a seizure, but also for epochs preceding seizures since the actual duration of the proictal phase prior to seizures can only be hypothesised. This leads to major challenges in the definition of a uniform framework for the development and comparison of new methods. Algorithms can be compared in their overall performance (e.g. time in false warning and sensitivity), but the identification of faulty behaviour is impossible considering the lack of a reliable ground truth. In the future, experimentally probing the cortical excitability via electric or transcranial magnetic stimulation \citep{Bauer2014, Freestone2011} could possibly provide hints whether the brain is in a state of increased seizure susceptibility.
 
%Next to the data-driven methods there have been attempts to find model-based approaches that take into account the physiology of the individual to understand the epileptic disease on the scale of single neurons, populations of neurons, and on a macroscopic scale from measurements of EEG \citep{Aarabi2014, Kuhlmann2015, Lehnertz2009, Rings2019, Wendling2005}. However, a statistical validation of such methods for performance evaluation in seizure prediction using long-term recordings is still pending. 

Finally, until data including reliable ground truth is available, we suggest to consider alternative approaches for data driven methods. By the way data driven models are currently trained, it is implicitly assumed that a deterministic preictal period precedes every seizure and that any other time is by definition interictal. Our findings support current studies that claim that this hypothesis might not hold for every patient and that probabilistic frameworks should be considered instead. Due to these developments, we propose the use of semi- or unsupervised trained models to acknowledge this fundamental change in the underlying hypothesis.
 
% xxx diese drei Abschnitte wiederholen sich und das schon vorher gesagte mehrmals. Ich würde "steps to improve kürzen. 
 
 %\begin{itemize}
 %	\item go beyond, unsupervised clustering of states, are there intrinsic difference between ii and pi? quantitative vs qualitative difference
 %	\item go beyond, dynamical trajectory of seizure development, adaptive definition of ii/pi, existence of a tipping point?
 %\end{itemize}

\section{Conclusion}

By comparing two substantially different seizure prediction algorithms on three  datasets we observed a remarkably strong coherence of correct but also of false predictions. As algorithms are predominantly sensitive to underlying changes in the data  the problem with apparently false predictions is unlikely to disappear by focusing on further optimizations of the algorithms for binary classification. 

In our opinion, we should instead focus on new working hypothesis in seizure prediction that follows a probabilistic rather than a deterministic approach. Considering a proictal state along with a clustering of the EEG data using unsupervised learning could be a promising approach.

\section*{Acknowledgement}

Ethical approval: The research related to human use complies with all the relevant national regulations, institutional policies and was performed in accordance with the tenets of the Helsinki Declaration, and has been approved by the authors' institutional review board or equivalent committee.

This work was supported by the European Regional Development Fund (ERDF), the Free State of Saxony (project number: 100320557), the Innovation Projects MedTech ALERT of Else Kröner Fresenius Center (EKFZ) for Digital Health of the TU Dresden and the University Hospital Carl Gustav Carus, and NHMRC project grant GNT1160815.

We thank Susanne Creutz for supporting the data acquisition and the Center for Information Services and High Performance Computing (ZIH) at TU Dresden for generous allocation of computing time.

\section*{Disclosure of Conflicts of Interest}
None of the authors has any conflict of interest to disclose.

\section*{Ethical Publication Statement}
We confirm that we have read the Journal's position on issues involved in ethical publication and affirm that this report is consistent with those guidelines.

\bibliographystyle{elsarticle-harv.bst}
\bibliography{bibo-epilepsy}

\begin{thebibliography}{38}
\expandafter\ifx\csname natexlab\endcsname\relax\def\natexlab#1{#1}\fi
\providecommand{\url}[1]{\texttt{#1}}
\providecommand{\href}[2]{#2}
\providecommand{\path}[1]{#1}
\providecommand{\DOIprefix}{doi:}
\providecommand{\ArXivprefix}{arXiv:}
\providecommand{\URLprefix}{URL: }
\providecommand{\Pubmedprefix}{pmid:}
\providecommand{\doi}[1]{\href{http://dx.doi.org/#1}{\path{#1}}}
\providecommand{\Pubmed}[1]{\href{pmid:#1}{\path{#1}}}
\providecommand{\bibinfo}[2]{#2}
\ifx\xfnm\relax \def\xfnm[#1]{\unskip,\space#1}\fi
%Type = Article
\bibitem[{Andrzejak et~al.(2003)Andrzejak, Kraskov, St{\"{o}}gbauer, Mormann
  and Kreuz}]{Andrzejak2003}
\bibinfo{author}{Andrzejak, R.G.}, \bibinfo{author}{Kraskov, A.},
  \bibinfo{author}{St{\"{o}}gbauer, H.}, \bibinfo{author}{Mormann, F.},
  \bibinfo{author}{Kreuz, T.}, \bibinfo{year}{2003}.
\newblock \bibinfo{title}{{Bivariate surrogate techniques: Necessity,
  strengths, and caveats}}.
\newblock \bibinfo{journal}{Physical Review E - Statistical Physics, Plasmas,
  Fluids, and Related Interdisciplinary Topics} \bibinfo{volume}{68}.
\newblock \DOIprefix\doi{10.1103/PhysRevE.68.066202}.
%Type = Article
\bibitem[{Bauer et~al.(2014)Bauer, Kalitzin, Zijlmans, Sander and
  Visser}]{Bauer2014}
\bibinfo{author}{Bauer, P.R.}, \bibinfo{author}{Kalitzin, S.},
  \bibinfo{author}{Zijlmans, M.}, \bibinfo{author}{Sander, J.W.},
  \bibinfo{author}{Visser, G.H.}, \bibinfo{year}{2014}.
\newblock \bibinfo{title}{{Cortical excitability as a potential clinical marker
  of epilepsy: A review of the clinical application of transcranial magnetic
  stimulation}}.
\newblock \bibinfo{journal}{Int. J. Neural Syst.} \bibinfo{volume}{24},
  \bibinfo{pages}{1430001}.
\newblock \URLprefix \url{www.worldscientific.com},
  \DOIprefix\doi{10.1142/S0129065714300010}.
%Type = Article
\bibitem[{Brinkmann et~al.(2016)Brinkmann, Wagenaar, Abbot, Adkins, Bosshard,
  Chen, Tieng, He, Mu{\~{n}}oz-Almaraz, Botella-Rocamora, Pardo,
  Zamora-Martinez, Hills, Wu, Korshunova, Cukierski, Vite, Patterson, Litt and
  Worrell}]{Brinkmann2016}
\bibinfo{author}{Brinkmann, B.H.}, \bibinfo{author}{Wagenaar, J.},
  \bibinfo{author}{Abbot, D.}, \bibinfo{author}{Adkins, P.},
  \bibinfo{author}{Bosshard, S.C.}, \bibinfo{author}{Chen, M.},
  \bibinfo{author}{Tieng, Q.M.}, \bibinfo{author}{He, J.},
  \bibinfo{author}{Mu{\~{n}}oz-Almaraz, F.J.},
  \bibinfo{author}{Botella-Rocamora, P.}, \bibinfo{author}{Pardo, J.},
  \bibinfo{author}{Zamora-Martinez, F.}, \bibinfo{author}{Hills, M.},
  \bibinfo{author}{Wu, W.}, \bibinfo{author}{Korshunova, I.},
  \bibinfo{author}{Cukierski, W.}, \bibinfo{author}{Vite, C.},
  \bibinfo{author}{Patterson, E.E.}, \bibinfo{author}{Litt, B.},
  \bibinfo{author}{Worrell, G.A.}, \bibinfo{year}{2016}.
\newblock \bibinfo{title}{{Crowdsourcing reproducible seizure forecasting in
  human and canine epilepsy}}.
\newblock \bibinfo{journal}{Brain} \bibinfo{volume}{139},
  \bibinfo{pages}{1713--1722}.
\newblock \URLprefix
  \url{https://academic.oup.com/brain/article-lookup/doi/10.1093/brain/aww045},
  \DOIprefix\doi{10.1093/brain/aww045}.
%Type = Article
\bibitem[{Coles et~al.(2013)Coles, Patterson, Sheffield, Mavoori, Higgins,
  Michael, Leyde, Cloyd, Litt, Vite and Worrell}]{Coles2013}
\bibinfo{author}{Coles, L.D.}, \bibinfo{author}{Patterson, E.E.},
  \bibinfo{author}{Sheffield, W.D.}, \bibinfo{author}{Mavoori, J.},
  \bibinfo{author}{Higgins, J.}, \bibinfo{author}{Michael, B.},
  \bibinfo{author}{Leyde, K.}, \bibinfo{author}{Cloyd, J.C.},
  \bibinfo{author}{Litt, B.}, \bibinfo{author}{Vite, C.},
  \bibinfo{author}{Worrell, G.A.}, \bibinfo{year}{2013}.
\newblock \bibinfo{title}{{Feasibility study of a caregiver seizure alert
  system in canine epilepsy}}.
\newblock \bibinfo{journal}{Epilepsy Res.} \bibinfo{volume}{106},
  \bibinfo{pages}{456--460}.
\newblock \URLprefix
  \url{https://www.sciencedirect.com/science/article/pii/S092012111300171X},
  \DOIprefix\doi{10.1016/J.EPLEPSYRES.2013.06.007}.
%Type = Article
\bibitem[{Cook et~al.(2013)Cook, O'Brien, Berkovic, Murphy, Morokoff, Fabinyi,
  D'Souza, Yerra, Archer, Litewka, Hosking, Lightfoot, Ruedebusch, Sheffield,
  Snyder, Leyde and Himes}]{Cook2013}
\bibinfo{author}{Cook, M.J.}, \bibinfo{author}{O'Brien, T.J.},
  \bibinfo{author}{Berkovic, S.F.}, \bibinfo{author}{Murphy, M.},
  \bibinfo{author}{Morokoff, A.}, \bibinfo{author}{Fabinyi, G.},
  \bibinfo{author}{D'Souza, W.}, \bibinfo{author}{Yerra, R.},
  \bibinfo{author}{Archer, J.}, \bibinfo{author}{Litewka, L.},
  \bibinfo{author}{Hosking, S.}, \bibinfo{author}{Lightfoot, P.},
  \bibinfo{author}{Ruedebusch, V.}, \bibinfo{author}{Sheffield, W.D.},
  \bibinfo{author}{Snyder, D.}, \bibinfo{author}{Leyde, K.},
  \bibinfo{author}{Himes, D.}, \bibinfo{year}{2013}.
\newblock \bibinfo{title}{{Prediction of seizure likelihood with a long-term,
  implanted seizure advisory system in patients with drug-resistant epilepsy: a
  first-in-man study}}.
\newblock \bibinfo{journal}{The Lancet Neurology} \bibinfo{volume}{12},
  \bibinfo{pages}{563--571}.
\newblock \URLprefix
  \url{https://linkinghub.elsevier.com/retrieve/pii/S1474442213700759},
  \DOIprefix\doi{10.1016/S1474-4422(13)70075-9}.
%Type = Article
\bibitem[{Davis et~al.(2011)Davis, Sturges, Vite, Ruedebusch, Worrell, Gardner,
  Leyde, Sheffield and Litt}]{Davis2011}
\bibinfo{author}{Davis, K.A.}, \bibinfo{author}{Sturges, B.K.},
  \bibinfo{author}{Vite, C.H.}, \bibinfo{author}{Ruedebusch, V.},
  \bibinfo{author}{Worrell, G.}, \bibinfo{author}{Gardner, A.B.},
  \bibinfo{author}{Leyde, K.}, \bibinfo{author}{Sheffield, W.D.},
  \bibinfo{author}{Litt, B.}, \bibinfo{year}{2011}.
\newblock \bibinfo{title}{{A novel implanted device to wirelessly record and
  analyze continuous intracranial canine EEG}}.
\newblock \bibinfo{journal}{Epilepsy Res.} \bibinfo{volume}{96},
  \bibinfo{pages}{116--122}.
\newblock \URLprefix
  \url{https://www.sciencedirect.com/science/article/pii/S0920121111001318?via{\%}3Dihub},
  \DOIprefix\doi{10.1016/J.EPLEPSYRES.2011.05.011}.
%Type = Article
\bibitem[{Direito et~al.(2017)Direito, Teixeira, Sales, Castelo-Branco and
  Dourado}]{Direito2017}
\bibinfo{author}{Direito, B.}, \bibinfo{author}{Teixeira, C.A.},
  \bibinfo{author}{Sales, F.}, \bibinfo{author}{Castelo-Branco, M.},
  \bibinfo{author}{Dourado, A.}, \bibinfo{year}{2017}.
\newblock \bibinfo{title}{{A Realistic Seizure Prediction Study Based on
  Multiclass SVM}}.
\newblock \bibinfo{journal}{Int. J. Neural Syst.} \bibinfo{volume}{27}.
\newblock \DOIprefix\doi{10.1142/S012906571750006X}.
%Type = Inproceedings
\bibitem[{{Eberlein} et~al.(2018){Eberlein}, {Hildebrand}, {Tetzlaff},
  {Hoffmann}, {Kuhlmann}, {Brinkmann} and {Müller}}]{Eberlein2018}
\bibinfo{author}{{Eberlein}, M.}, \bibinfo{author}{{Hildebrand}, R.},
  \bibinfo{author}{{Tetzlaff}, R.}, \bibinfo{author}{{Hoffmann}, N.},
  \bibinfo{author}{{Kuhlmann}, L.}, \bibinfo{author}{{Brinkmann}, B.},
  \bibinfo{author}{{Müller}, J.}, \bibinfo{year}{2018}.
\newblock \bibinfo{title}{{Convolutional Neural Networks for Epileptic Seizure
  Prediction}}, in: \bibinfo{booktitle}{Proc. IEEE Int. Conf. Bioinformatics
  and Biomedicine (BIBM)}, pp. \bibinfo{pages}{2577--2582}.
\newblock \DOIprefix\doi{10.1109/BIBM.2018.8621225}.
%Type = Article
\bibitem[{Eberlein et~al.(2019)Eberlein, M{\"u}ller, Yang, Walz, Schreiber,
  Tetzlaff, Creutz, Uckermann and Leonhardt}]{Eberlein2019}
\bibinfo{author}{Eberlein, M.}, \bibinfo{author}{M{\"u}ller, J.},
  \bibinfo{author}{Yang, H.}, \bibinfo{author}{Walz, S.},
  \bibinfo{author}{Schreiber, J.}, \bibinfo{author}{Tetzlaff, R.},
  \bibinfo{author}{Creutz, S.}, \bibinfo{author}{Uckermann, O.},
  \bibinfo{author}{Leonhardt, G.}, \bibinfo{year}{2019}.
\newblock \bibinfo{title}{Evaluation of machine learning methods for seizure
  prediction in epilepsy}.
\newblock \bibinfo{journal}{Current Directions in Biomedical Engineering}
  \bibinfo{volume}{5}, \bibinfo{pages}{109--112}.
%Type = Article
\bibitem[{Freestone et~al.(2017)Freestone, Karoly and Cook}]{Freestone2017}
\bibinfo{author}{Freestone, D.R.}, \bibinfo{author}{Karoly, P.J.},
  \bibinfo{author}{Cook, M.J.}, \bibinfo{year}{2017}.
\newblock \bibinfo{title}{{A forward-looking review of seizure prediction}}.
\newblock \bibinfo{journal}{Curr. Opin. Neurol.} \bibinfo{volume}{30},
  \bibinfo{pages}{167--173}.
\newblock \URLprefix \url{www.co-neurology.com
  http://insights.ovid.com/crossref?an=00019052-201704000-00009},
  \DOIprefix\doi{10.1097/WCO.0000000000000429}.
%Type = Article
\bibitem[{Freestone et~al.(2011)Freestone, Kuhlmann, Grayden, Burkitt, Lai,
  Nelson, Vogrin, Murphy, D'Souza, Badawy, Nesic and Cook}]{Freestone2011}
\bibinfo{author}{Freestone, D.R.}, \bibinfo{author}{Kuhlmann, L.},
  \bibinfo{author}{Grayden, D.B.}, \bibinfo{author}{Burkitt, A.N.},
  \bibinfo{author}{Lai, A.}, \bibinfo{author}{Nelson, T.S.},
  \bibinfo{author}{Vogrin, S.}, \bibinfo{author}{Murphy, M.},
  \bibinfo{author}{D'Souza, W.}, \bibinfo{author}{Badawy, R.},
  \bibinfo{author}{Nesic, D.}, \bibinfo{author}{Cook, M.J.},
  \bibinfo{year}{2011}.
\newblock \bibinfo{title}{{Electrical probing of cortical excitability in
  patients with epilepsy}}.
\newblock \bibinfo{journal}{Epilepsy and Behavior} \bibinfo{volume}{22},
  \bibinfo{pages}{S110--S118}.
\newblock \DOIprefix\doi{10.1016/j.yebeh.2011.09.005}.
%Type = Article
\bibitem[{Gadhoumi et~al.(2016)Gadhoumi, Lina, Mormann and
  Gotman}]{Gadhoumi2016}
\bibinfo{author}{Gadhoumi, K.}, \bibinfo{author}{Lina, J.M.},
  \bibinfo{author}{Mormann, F.}, \bibinfo{author}{Gotman, J.},
  \bibinfo{year}{2016}.
\newblock \bibinfo{title}{{Seizure prediction for therapeutic devices: A
  review}}.
\newblock \bibinfo{journal}{J. Neurosci. Methods} \bibinfo{volume}{260},
  \bibinfo{pages}{270--282}.
\newblock \URLprefix
  \url{https://www.sciencedirect.com/science/article/pii/S0165027015002277},
  \DOIprefix\doi{10.1016/j.jneumeth.2015.06.010}.
%Type = Article
\bibitem[{Ghaderyan et~al.(2014)Ghaderyan, Abbasi and Sedaaghi}]{Ghaderyan2014}
\bibinfo{author}{Ghaderyan, P.}, \bibinfo{author}{Abbasi, A.},
  \bibinfo{author}{Sedaaghi, M.H.}, \bibinfo{year}{2014}.
\newblock \bibinfo{title}{{An efficient seizure prediction method using
  KNN-based undersampling and linear frequency measures}}.
\newblock \bibinfo{journal}{J. Neurosci. Methods} \bibinfo{volume}{232},
  \bibinfo{pages}{134--142}.
\newblock \DOIprefix\doi{10.1016/j.jneumeth.2014.05.019}.
%Type = Article
\bibitem[{Hanley and J.(1982)}]{Hanley1982}
\bibinfo{author}{Hanley, J.A.}, \bibinfo{author}{J., M.B.},
  \bibinfo{year}{1982}.
\newblock \bibinfo{title}{{The Meaning and Use of the Area under a Receiver
  Operating Characteristic (ROC) Curve}}.
\newblock \bibinfo{journal}{Radiology} \bibinfo{volume}{143},
  \bibinfo{pages}{29--36}.
%Type = Article
\bibitem[{Howbert et~al.(2014)Howbert, Patterson, Stead, Brinkmann, Vasoli,
  Crepeau, Vite, Sturges, Ruedebusch, Mavoori, Leyde, Sheffield, Litt and
  Worrell}]{Howbert2014}
\bibinfo{author}{Howbert, J.J.}, \bibinfo{author}{Patterson, E.E.},
  \bibinfo{author}{Stead, S.M.}, \bibinfo{author}{Brinkmann, B.},
  \bibinfo{author}{Vasoli, V.}, \bibinfo{author}{Crepeau, D.},
  \bibinfo{author}{Vite, C.H.}, \bibinfo{author}{Sturges, B.},
  \bibinfo{author}{Ruedebusch, V.}, \bibinfo{author}{Mavoori, J.},
  \bibinfo{author}{Leyde, K.}, \bibinfo{author}{Sheffield, W.D.},
  \bibinfo{author}{Litt, B.}, \bibinfo{author}{Worrell, G.A.},
  \bibinfo{year}{2014}.
\newblock \bibinfo{title}{{Forecasting seizures in dogs with naturally
  occurring epilepsy}}.
\newblock \bibinfo{journal}{PLoS One} \bibinfo{volume}{9},
  \bibinfo{pages}{e81920}.
\newblock \URLprefix \url{https://dx.plos.org/10.1371/journal.pone.0081920},
  \DOIprefix\doi{10.1371/journal.pone.0081920}.
%Type = Article
\bibitem[{Kalitzin et~al.(2011)Kalitzin, Koppert, Petkov, Velis and
  da~Silva}]{Kalitzin2011}
\bibinfo{author}{Kalitzin, S.}, \bibinfo{author}{Koppert, M.},
  \bibinfo{author}{Petkov, G.}, \bibinfo{author}{Velis, D.},
  \bibinfo{author}{da~Silva, F.L.}, \bibinfo{year}{2011}.
\newblock \bibinfo{title}{{Computational model prospective on the observation
  of proictal states in epileptic neuronal systems}}.
\newblock \bibinfo{journal}{Epilepsy {\&} Behavior} \bibinfo{volume}{22},
  \bibinfo{pages}{S102--S109}.
\newblock \URLprefix
  \url{https://linkinghub.elsevier.com/retrieve/pii/S1525505011004811},
  \DOIprefix\doi{10.1016/j.yebeh.2011.08.017}.
%Type = Article
\bibitem[{Karoly et~al.(2017)Karoly, Ung, Grayden, Kuhlmann, Leyde, Cook and
  Freestone}]{Karoly2017}
\bibinfo{author}{Karoly, P.J.}, \bibinfo{author}{Ung, H.},
  \bibinfo{author}{Grayden, D.B.}, \bibinfo{author}{Kuhlmann, L.},
  \bibinfo{author}{Leyde, K.}, \bibinfo{author}{Cook, M.J.},
  \bibinfo{author}{Freestone, D.R.}, \bibinfo{year}{2017}.
\newblock \bibinfo{title}{{The circadian profile of epilepsy improves seizure
  forecasting}}.
\newblock \bibinfo{journal}{Brain} \bibinfo{volume}{140},
  \bibinfo{pages}{2169--2182}.
\newblock \URLprefix
  \url{https://academic.oup.com/brain/article/140/8/2169/4032453},
  \DOIprefix\doi{10.1093/brain/awx173}.
%Type = Article
\bibitem[{Korshunova et~al.(2018)Korshunova, Kindermans, Degrave, Verhoeven,
  Brinkmann and Dambre}]{Korshunova2018}
\bibinfo{author}{Korshunova, I.}, \bibinfo{author}{Kindermans, P.J.},
  \bibinfo{author}{Degrave, J.}, \bibinfo{author}{Verhoeven, T.},
  \bibinfo{author}{Brinkmann, B.H.}, \bibinfo{author}{Dambre, J.},
  \bibinfo{year}{2018}.
\newblock \bibinfo{title}{{Towards Improved Design and Evaluation of Epileptic
  Seizure Predictors}}.
\newblock \bibinfo{journal}{IEEE Trans. Biomed. Eng.}
  \DOIprefix\doi{10.1109/TBME.2017.2700086}.
%Type = Article
\bibitem[{Kuhlmann et~al.(2010)Kuhlmann, Freestone, Lai, Burkitt, Fuller,
  Grayden, Seiderer, Vogrin, Mareels and Cook}]{Kuhlmann2010}
\bibinfo{author}{Kuhlmann, L.}, \bibinfo{author}{Freestone, D.},
  \bibinfo{author}{Lai, A.}, \bibinfo{author}{Burkitt, A.N.},
  \bibinfo{author}{Fuller, K.}, \bibinfo{author}{Grayden, D.B.},
  \bibinfo{author}{Seiderer, L.}, \bibinfo{author}{Vogrin, S.},
  \bibinfo{author}{Mareels, I.M.}, \bibinfo{author}{Cook, M.J.},
  \bibinfo{year}{2010}.
\newblock \bibinfo{title}{{Patient-specific bivariate-synchrony-based seizure
  prediction for short prediction horizons}}.
\newblock \bibinfo{journal}{Epilepsy Res.} \bibinfo{volume}{91},
  \bibinfo{pages}{214--231}.
\newblock \DOIprefix\doi{10.1016/j.eplepsyres.2010.07.014}.
%Type = Article
\bibitem[{Kuhlmann et~al.(2018a)Kuhlmann, Karoly, Freestone, Brinkmann, Temko,
  Barachant, Li, Titericz, Lang, Lavery, Roman, Broadhead, Dobson, Jones, Tang,
  Ivanenko, Panichev, Proix, N{\'{a}}hl{\'{i}}k, Grunberg, Reuben, Worrell,
  Litt, Liley, Grayden and Cook}]{Kuhlmann2018a}
\bibinfo{author}{Kuhlmann, L.}, \bibinfo{author}{Karoly, P.},
  \bibinfo{author}{Freestone, D.R.}, \bibinfo{author}{Brinkmann, B.H.},
  \bibinfo{author}{Temko, A.}, \bibinfo{author}{Barachant, A.},
  \bibinfo{author}{Li, F.}, \bibinfo{author}{Titericz, G.},
  \bibinfo{author}{Lang, B.W.}, \bibinfo{author}{Lavery, D.},
  \bibinfo{author}{Roman, K.}, \bibinfo{author}{Broadhead, D.},
  \bibinfo{author}{Dobson, S.}, \bibinfo{author}{Jones, G.},
  \bibinfo{author}{Tang, Q.}, \bibinfo{author}{Ivanenko, I.},
  \bibinfo{author}{Panichev, O.}, \bibinfo{author}{Proix, T.},
  \bibinfo{author}{N{\'{a}}hl{\'{i}}k, M.}, \bibinfo{author}{Grunberg, D.B.},
  \bibinfo{author}{Reuben, C.}, \bibinfo{author}{Worrell, G.},
  \bibinfo{author}{Litt, B.}, \bibinfo{author}{Liley, D.T.J.},
  \bibinfo{author}{Grayden, D.B.}, \bibinfo{author}{Cook, M.J.},
  \bibinfo{year}{2018}a.
\newblock \bibinfo{title}{{Epilepsyecosystem.org: crowd-sourcing reproducible
  seizure prediction with long-term human intracranial EEG}}.
\newblock \bibinfo{journal}{Brain} \bibinfo{volume}{141},
  \bibinfo{pages}{2619--2630}.
\newblock \URLprefix
  \url{https://academic.oup.com/brain/advance-article/doi/10.1093/brain/awy210/5066003},
  \DOIprefix\doi{10.1093/brain/awy210}.
%Type = Article
\bibitem[{Kuhlmann et~al.(2018b)Kuhlmann, Lehnertz, Richardson, Schelter and
  Zaveri}]{Kuhlmann2018}
\bibinfo{author}{Kuhlmann, L.}, \bibinfo{author}{Lehnertz, K.},
  \bibinfo{author}{Richardson, M.P.}, \bibinfo{author}{Schelter, B.},
  \bibinfo{author}{Zaveri, H.P.}, \bibinfo{year}{2018}b.
\newblock \bibinfo{title}{Seizure prediction - ready for a new era.}
\newblock \bibinfo{journal}{Nature reviews. Neurology} \bibinfo{volume}{14},
  \bibinfo{pages}{618--630}.
\newblock \DOIprefix\doi{10.1038/s41582-018-0055-2}.
%Type = Inproceedings
\bibitem[{Ma et~al.(2018)Ma, Qiu, Zhang, Lian and He}]{Ma2018}
\bibinfo{author}{Ma, X.}, \bibinfo{author}{Qiu, S.}, \bibinfo{author}{Zhang,
  Y.}, \bibinfo{author}{Lian, X.}, \bibinfo{author}{He, H.},
  \bibinfo{year}{2018}.
\newblock \bibinfo{title}{Predicting epileptic seizures from intracranial eeg
  using lstm-based multi-task learning}, in: \bibinfo{editor}{Lai, J.H.},
  \bibinfo{editor}{Liu, C.L.}, \bibinfo{editor}{Chen, X.},
  \bibinfo{editor}{Zhou, J.}, \bibinfo{editor}{Tan, T.},
  \bibinfo{editor}{Zheng, N.}, \bibinfo{editor}{Zha, H.} (Eds.),
  \bibinfo{booktitle}{Pattern Recognition and Computer Vision},
  \bibinfo{publisher}{Springer International Publishing},
  \bibinfo{address}{Cham}. pp. \bibinfo{pages}{157--167}.
%Type = Article
\bibitem[{Meisel et~al.(2015)Meisel, Schulze-Bonhage, Freestone, Cook,
  Achermann and Plenz}]{Meisel2015}
\bibinfo{author}{Meisel, C.}, \bibinfo{author}{Schulze-Bonhage, A.},
  \bibinfo{author}{Freestone, D.}, \bibinfo{author}{Cook, M.J.},
  \bibinfo{author}{Achermann, P.}, \bibinfo{author}{Plenz, D.},
  \bibinfo{year}{2015}.
\newblock \bibinfo{title}{{Intrinsic excitability measures track antiepileptic
  drug action and uncover increasing/decreasing excitability over the
  wake/sleep cycle}}.
\newblock \bibinfo{journal}{Proc Natl Acad Sci USA} \bibinfo{volume}{112},
  \bibinfo{pages}{14694--14699}.
\newblock \URLprefix \url{www.pnas.org/cgi/doi/10.1073/pnas.1513716112},
  \DOIprefix\doi{10.1073/pnas.1513716112}.
%Type = Article
\bibitem[{Mormann and Andrzejak(2016)}]{Mormann2016}
\bibinfo{author}{Mormann, F.}, \bibinfo{author}{Andrzejak, R.G.},
  \bibinfo{year}{2016}.
\newblock \bibinfo{title}{{Seizure prediction: making mileage on the long and
  winding road}}.
\newblock \bibinfo{journal}{Brain} \bibinfo{volume}{139},
  \bibinfo{pages}{1625--1627}.
\newblock \URLprefix \url{https://doi.org/10.1093/brain/aww091},
  \DOIprefix\doi{10.1093/brain/aww091}.
%Type = Article
\bibitem[{Mormann et~al.(2007)Mormann, Andrzejak, Elger and
  Lehnertz}]{Mormann2007}
\bibinfo{author}{Mormann, F.}, \bibinfo{author}{Andrzejak, R.G.},
  \bibinfo{author}{Elger, C.E.}, \bibinfo{author}{Lehnertz, K.},
  \bibinfo{year}{2007}.
\newblock \bibinfo{title}{Seizure prediction: the long and winding road}.
\newblock \bibinfo{journal}{Brain} \bibinfo{volume}{130},
  \bibinfo{pages}{314--333}.
%Type = Article
\bibitem[{Mormann et~al.(2003)Mormann, Andrzejak, Kreuz, Rieke, David, Elger
  and Lehnertz}]{Mormann2003}
\bibinfo{author}{Mormann, F.}, \bibinfo{author}{Andrzejak, R.G.},
  \bibinfo{author}{Kreuz, T.}, \bibinfo{author}{Rieke, C.},
  \bibinfo{author}{David, P.}, \bibinfo{author}{Elger, C.E.},
  \bibinfo{author}{Lehnertz, K.}, \bibinfo{year}{2003}.
\newblock \bibinfo{title}{{Automated detection of a preseizure state based on a
  decrease in synchronization in intracranial electroencephalogram recordings
  from epilepsy patients}}.
\newblock \bibinfo{journal}{Physical Review E - Statistical Physics, Plasmas,
  Fluids, and Related Interdisciplinary Topics} \bibinfo{volume}{67},
  \bibinfo{pages}{10}.
\newblock \DOIprefix\doi{10.1103/PhysRevE.67.021912}.
%Type = Article
\bibitem[{Mormann et~al.(2005)Mormann, Kreuz, Rieke, Andrzejak, Kraskov, David,
  Elger and Lehnertz}]{Mormann2005}
\bibinfo{author}{Mormann, F.}, \bibinfo{author}{Kreuz, T.},
  \bibinfo{author}{Rieke, C.}, \bibinfo{author}{Andrzejak, R.G.},
  \bibinfo{author}{Kraskov, A.}, \bibinfo{author}{David, P.},
  \bibinfo{author}{Elger, C.E.}, \bibinfo{author}{Lehnertz, K.},
  \bibinfo{year}{2005}.
\newblock \bibinfo{title}{{On the predictability of epileptic seizures}}.
\newblock \bibinfo{journal}{Clin. Neurophysiol.} \bibinfo{volume}{116},
  \bibinfo{pages}{569--587}.
\newblock \URLprefix
  \url{https://www.sciencedirect.com/science/article/pii/S1388245704004638},
  \DOIprefix\doi{10.1016/J.CLINPH.2004.08.025}.
%Type = Article
\bibitem[{Nejedly et~al.(2019)Nejedly, Kremen, Sladky, Nasseri, Guragain,
  Klimes, Cimbalnik, Varatharajah, Brinkmann and Worrell}]{Nejedly2019}
\bibinfo{author}{Nejedly, P.}, \bibinfo{author}{Kremen, V.},
  \bibinfo{author}{Sladky, V.}, \bibinfo{author}{Nasseri, M.},
  \bibinfo{author}{Guragain, H.}, \bibinfo{author}{Klimes, P.},
  \bibinfo{author}{Cimbalnik, J.}, \bibinfo{author}{Varatharajah, Y.},
  \bibinfo{author}{Brinkmann, B.H.}, \bibinfo{author}{Worrell, G.A.},
  \bibinfo{year}{2019}.
\newblock \bibinfo{title}{{Deep-learning for seizure forecasting in canines
  with epilepsy}}.
\newblock \bibinfo{journal}{J. Neural Eng.} \bibinfo{volume}{16},
  \bibinfo{pages}{036031}.
\newblock \URLprefix
  \url{https://iopscience.iop.org/article/10.1088/1741-2552/ab172d},
  \DOIprefix\doi{10.1088/1741-2552/ab172d}.
%Type = Article
\bibitem[{Proix et~al.(2019)Proix, Truccolo, Leguia, King-Stephens, Rao and
  Baud}]{Proix2019}
\bibinfo{author}{Proix, T.}, \bibinfo{author}{Truccolo, W.},
  \bibinfo{author}{Leguia, M.G.}, \bibinfo{author}{King-Stephens, D.},
  \bibinfo{author}{Rao, V.R.}, \bibinfo{author}{Baud, M.O.},
  \bibinfo{year}{2019}.
\newblock \bibinfo{title}{{Forecasting Seizure Risk over Days}}.
\newblock \bibinfo{journal}{MedRxiv preprint} \URLprefix
  \url{http://dx.doi.org/10.1101/19008086}, \DOIprefix\doi{10.1101/19008086}.
%Type = Article
\bibitem[{Reuben et~al.(2019)Reuben, Karoly, Freestone, Temko, Barachant, Li,
  Titericz, Lang, Lavery, Roman, Broadhead, Jones, Tang, Ivanenko, Panichev,
  Proix, N{\'{a}}hl{\'{i}}k, Grunberg, Grayden, Cook and Kuhlmann}]{Reuben2019}
\bibinfo{author}{Reuben, C.}, \bibinfo{author}{Karoly, P.},
  \bibinfo{author}{Freestone, D.R.}, \bibinfo{author}{Temko, A.},
  \bibinfo{author}{Barachant, A.}, \bibinfo{author}{Li, F.},
  \bibinfo{author}{Titericz, G.}, \bibinfo{author}{Lang, B.W.},
  \bibinfo{author}{Lavery, D.}, \bibinfo{author}{Roman, K.},
  \bibinfo{author}{Broadhead, D.}, \bibinfo{author}{Jones, G.},
  \bibinfo{author}{Tang, Q.}, \bibinfo{author}{Ivanenko, I.},
  \bibinfo{author}{Panichev, O.}, \bibinfo{author}{Proix, T.},
  \bibinfo{author}{N{\'{a}}hl{\'{i}}k, M.}, \bibinfo{author}{Grunberg, D.B.},
  \bibinfo{author}{Grayden, D.B.}, \bibinfo{author}{Cook, M.J.},
  \bibinfo{author}{Kuhlmann, L.}, \bibinfo{year}{2019}.
\newblock \bibinfo{title}{{Ensembling crowdsourced seizure prediction
  algorithms using long‐term human intracranial EEG}}.
\newblock \bibinfo{journal}{Epilepsia} , \bibinfo{pages}{epi.16418}\URLprefix
  \url{https://onlinelibrary.wiley.com/doi/abs/10.1111/epi.16418},
  \DOIprefix\doi{10.1111/epi.16418}.
%Type = Article
\bibitem[{Senger and Tetzlaff(2016)}]{Senger2016}
\bibinfo{author}{Senger, V.}, \bibinfo{author}{Tetzlaff, R.},
  \bibinfo{year}{2016}.
\newblock \bibinfo{title}{{New Signal Processing Methods for the Development of
  Seizure Warning Devices in Epilepsy}}.
\newblock \bibinfo{journal}{IEEE Trans. Circuits Syst. I} \bibinfo{volume}{63},
  \bibinfo{pages}{609--616}.
\newblock \DOIprefix\doi{10.1109/TCSI.2016.2553278}.
%Type = Article
\bibitem[{Snyder et~al.(2008)Snyder, Echauz, Grimes and Litt}]{Snyder2008}
\bibinfo{author}{Snyder, D.E.}, \bibinfo{author}{Echauz, J.},
  \bibinfo{author}{Grimes, D.B.}, \bibinfo{author}{Litt, B.},
  \bibinfo{year}{2008}.
\newblock \bibinfo{title}{{The statistics of a practical seizure warning
  system}}.
\newblock \bibinfo{journal}{J. Neural Eng.} \bibinfo{volume}{5},
  \bibinfo{pages}{392--401}.
\newblock \URLprefix
  \url{http://stacks.iop.org/1741-2552/5/i=4/a=004?key=crossref.c4740c42de2f75b6e8e598babbe332bd},
  \DOIprefix\doi{10.1088/1741-2560/5/4/004}.
%Type = Article
\bibitem[{Stirling et~al.(2020)Stirling, Cook, Grayden and
  Karoly}]{Stirling2020}
\bibinfo{author}{Stirling, R.E.}, \bibinfo{author}{Cook, M.J.},
  \bibinfo{author}{Grayden, D.B.}, \bibinfo{author}{Karoly, P.J.},
  \bibinfo{year}{2020}.
\newblock \bibinfo{title}{{Seizure forecasting and cyclic control of
  seizures}}.
\newblock \bibinfo{journal}{Epilepsia} , \bibinfo{pages}{epi.16541}\URLprefix
  \url{https://onlinelibrary.wiley.com/doi/abs/10.1111/epi.16541},
  \DOIprefix\doi{10.1111/epi.16541}.
%Type = Article
\bibitem[{Tetzlaff and Senger(2012)}]{Tetzlaff2012}
\bibinfo{author}{Tetzlaff, R.}, \bibinfo{author}{Senger, V.},
  \bibinfo{year}{2012}.
\newblock \bibinfo{title}{The seizure prediction problem in epilepsy: Cellular
  nonlinear networks}.
\newblock \bibinfo{journal}{Circuits and Systems Magazine, IEEE}
  \bibinfo{volume}{12}, \bibinfo{pages}{8--20}.
%Type = Article
\bibitem[{Tsiouris et~al.(2018)Tsiouris, Pezoulas, Zervakis, Konitsiotis,
  Koutsouris and Fotiadis}]{Tsiouris2018}
\bibinfo{author}{Tsiouris, K.M.}, \bibinfo{author}{Pezoulas, V.C.},
  \bibinfo{author}{Zervakis, M.}, \bibinfo{author}{Konitsiotis, S.},
  \bibinfo{author}{Koutsouris, D.D.}, \bibinfo{author}{Fotiadis, D.I.},
  \bibinfo{year}{2018}.
\newblock \bibinfo{title}{{A Long Short-Term Memory deep learning network for
  the prediction of epileptic seizures using EEG signals}}.
\newblock \bibinfo{journal}{Comput. Biol. Med.} \bibinfo{volume}{99},
  \bibinfo{pages}{24--37}.
\newblock \URLprefix
  \url{https://www.sciencedirect.com/science/article/pii/S001048251830132X?via{\%}3Dihub},
  \DOIprefix\doi{10.1016/J.COMPBIOMED.2018.05.019}.
%Type = Article
\bibitem[{Ung et~al.(2016)Ung, Davis, Wulsin, Wagenaar, Fox, McDonnell,
  Patterson, Vite, Worrell and Litt}]{Ung2016}
\bibinfo{author}{Ung, H.}, \bibinfo{author}{Davis, K.A.},
  \bibinfo{author}{Wulsin, D.}, \bibinfo{author}{Wagenaar, J.},
  \bibinfo{author}{Fox, E.}, \bibinfo{author}{McDonnell, J.J.},
  \bibinfo{author}{Patterson, N.}, \bibinfo{author}{Vite, C.H.},
  \bibinfo{author}{Worrell, G.}, \bibinfo{author}{Litt, B.},
  \bibinfo{year}{2016}.
\newblock \bibinfo{title}{{Temporal behavior of seizures and interictal bursts
  in prolonged intracranial recordings from epileptic canines}}.
\newblock \bibinfo{journal}{Epilepsia} \bibinfo{volume}{57},
  \bibinfo{pages}{1949--1957}.
\newblock \URLprefix \url{http://doi.wiley.com/10.1111/epi.13591},
  \DOIprefix\doi{10.1111/epi.13591}.
%Type = Article
\bibitem[{Wagenaar et~al.(2015)Wagenaar, Worrell, Ives, Matthias, Litt and
  Schulze-Bonhage}]{Wagenaar2015}
\bibinfo{author}{Wagenaar, J.B.}, \bibinfo{author}{Worrell, G.A.},
  \bibinfo{author}{Ives, Z.}, \bibinfo{author}{Matthias, D.},
  \bibinfo{author}{Litt, B.}, \bibinfo{author}{Schulze-Bonhage, A.},
  \bibinfo{year}{2015}.
\newblock \bibinfo{title}{{Collaborating and sharing data in epilepsy
  research}}.
\newblock \bibinfo{journal}{J. Clin. Neurophysiol.} \bibinfo{volume}{32},
  \bibinfo{pages}{235--239}.
\newblock \URLprefix \url{/pmc/articles/PMC4455031/?report=abstract
  https://www.ncbi.nlm.nih.gov/pmc/articles/PMC4455031/},
  \DOIprefix\doi{10.1097/WNP.0000000000000159}.
%Type = Article
\bibitem[{Zheng et~al.(2014)Zheng, Wang, Li, Bao and Wang}]{Zheng2014}
\bibinfo{author}{Zheng, Y.}, \bibinfo{author}{Wang, G.}, \bibinfo{author}{Li,
  K.}, \bibinfo{author}{Bao, G.}, \bibinfo{author}{Wang, J.},
  \bibinfo{year}{2014}.
\newblock \bibinfo{title}{{Epileptic seizure prediction using phase
  synchronization based on bivariate empirical mode decomposition}}.
\newblock \bibinfo{journal}{Clin. Neurophysiol.} \bibinfo{volume}{125},
  \bibinfo{pages}{1104--1111}.
\newblock \DOIprefix\doi{10.1016/j.clinph.2013.09.047}.

\end{thebibliography}

\newpage

\section*{Supplementary Material}
\subsection*{Prediction Plots}

Prediction over time for the test set of all 5 patients of dataset 1

\begin{figure}[H]
	\centering
	\subfigure[Patient A]{ \includegraphics[width=0.90\columnwidth]{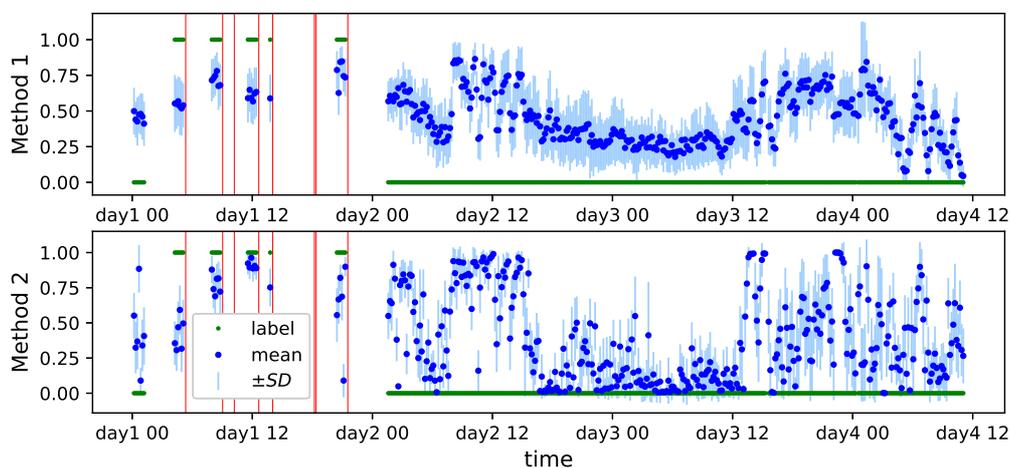}}
	\subfigure[Patient B]{ \includegraphics[width=0.90\columnwidth]{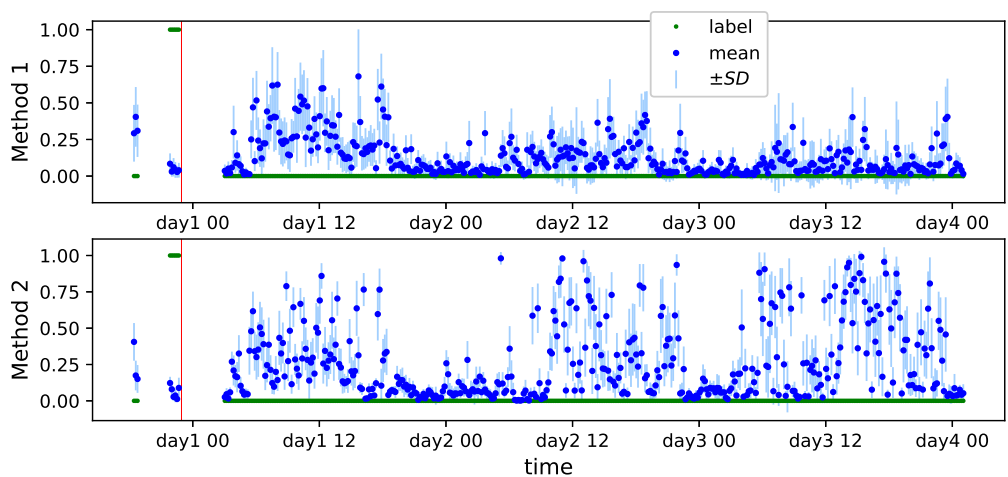}}
	\caption{Predictions for patients A, B of dataset 1. Seizure onset is marked as red line.}	
	\label{fig:ukdd-prb-part1}
\end{figure}

\begin{figure}[H]
	\centering
	\subfigure[Patient C]{ \includegraphics[width=0.90\columnwidth]{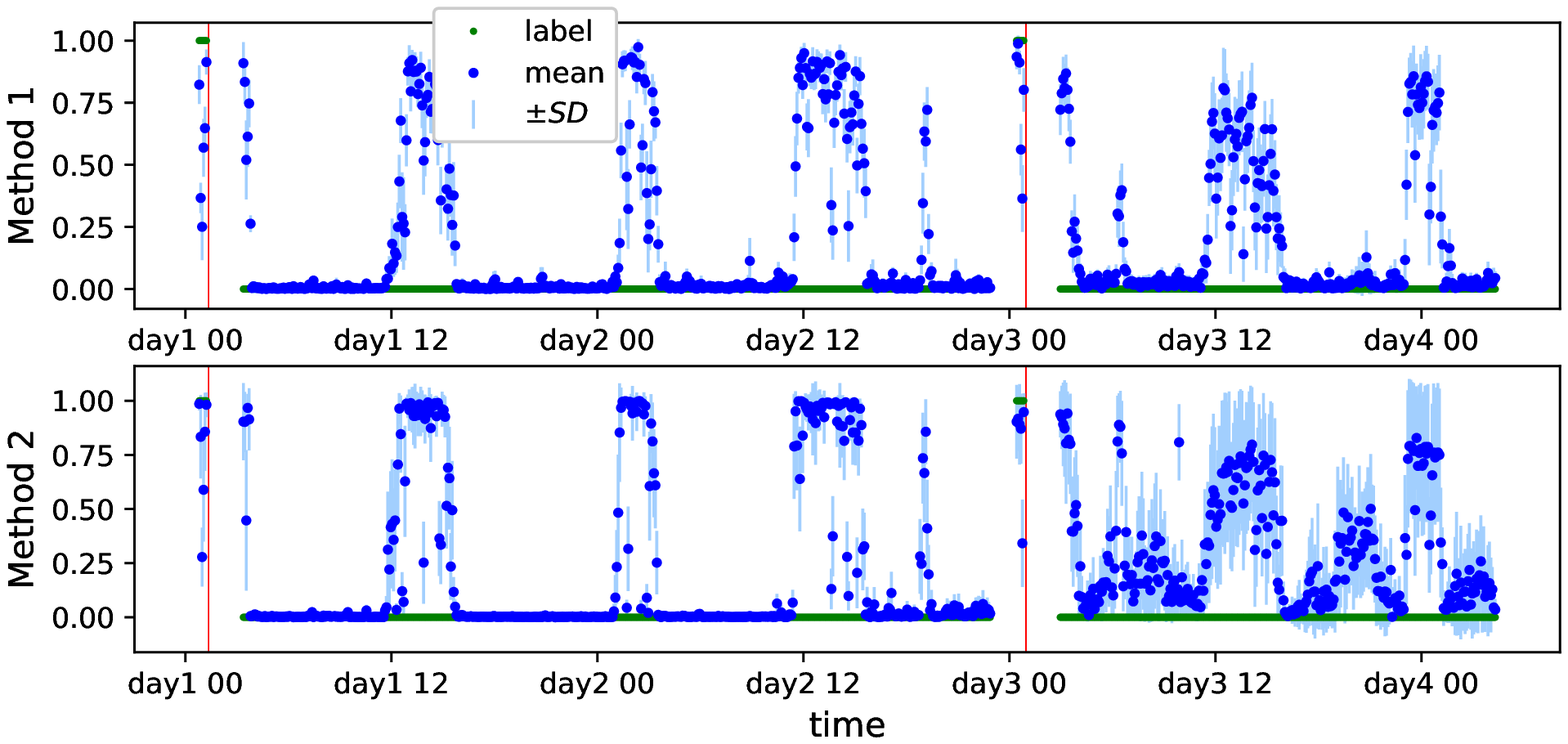}}
	\subfigure[Patient D]{ \includegraphics[width=0.90\columnwidth]{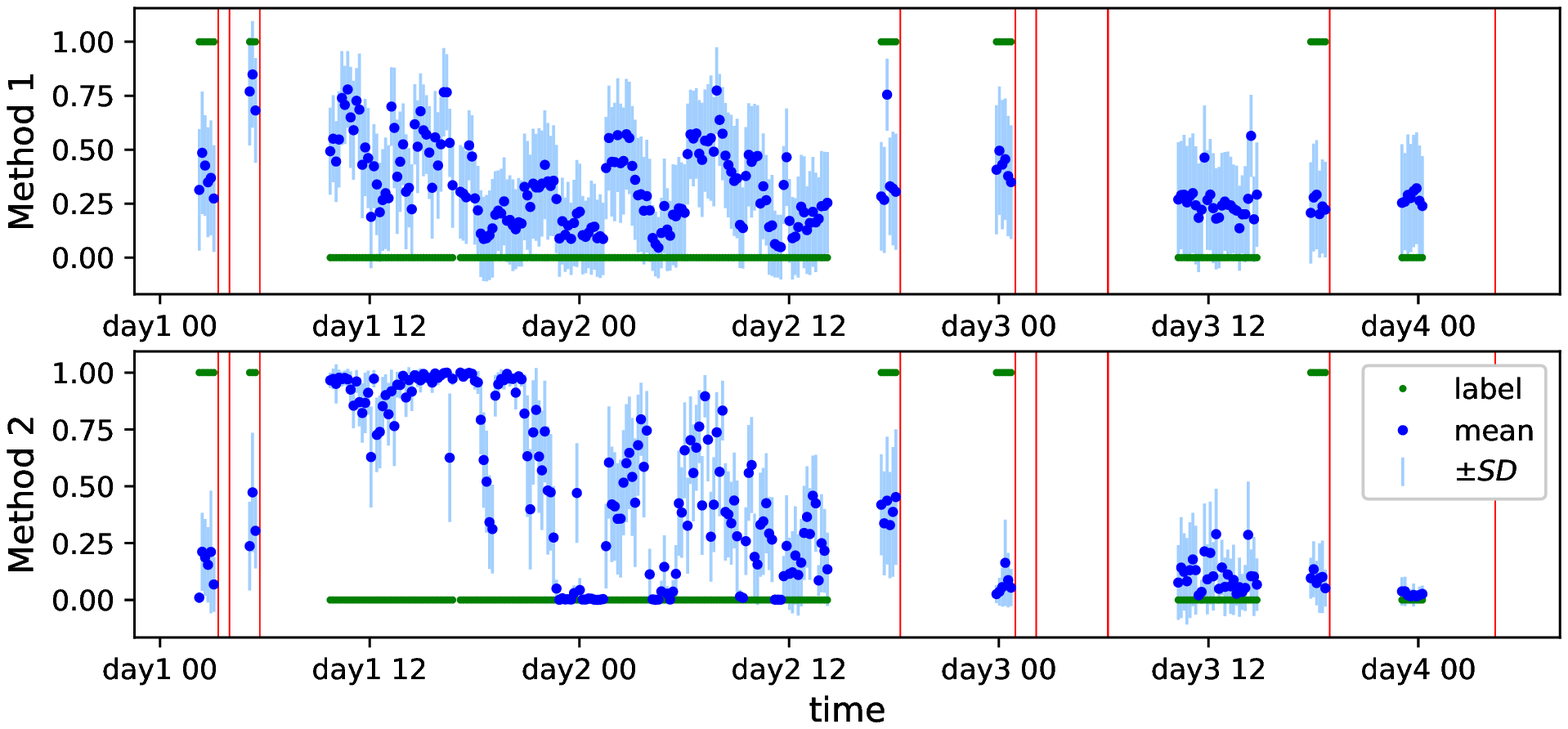}}
	\subfigure[Patient E]{ \includegraphics[width=0.90\columnwidth]{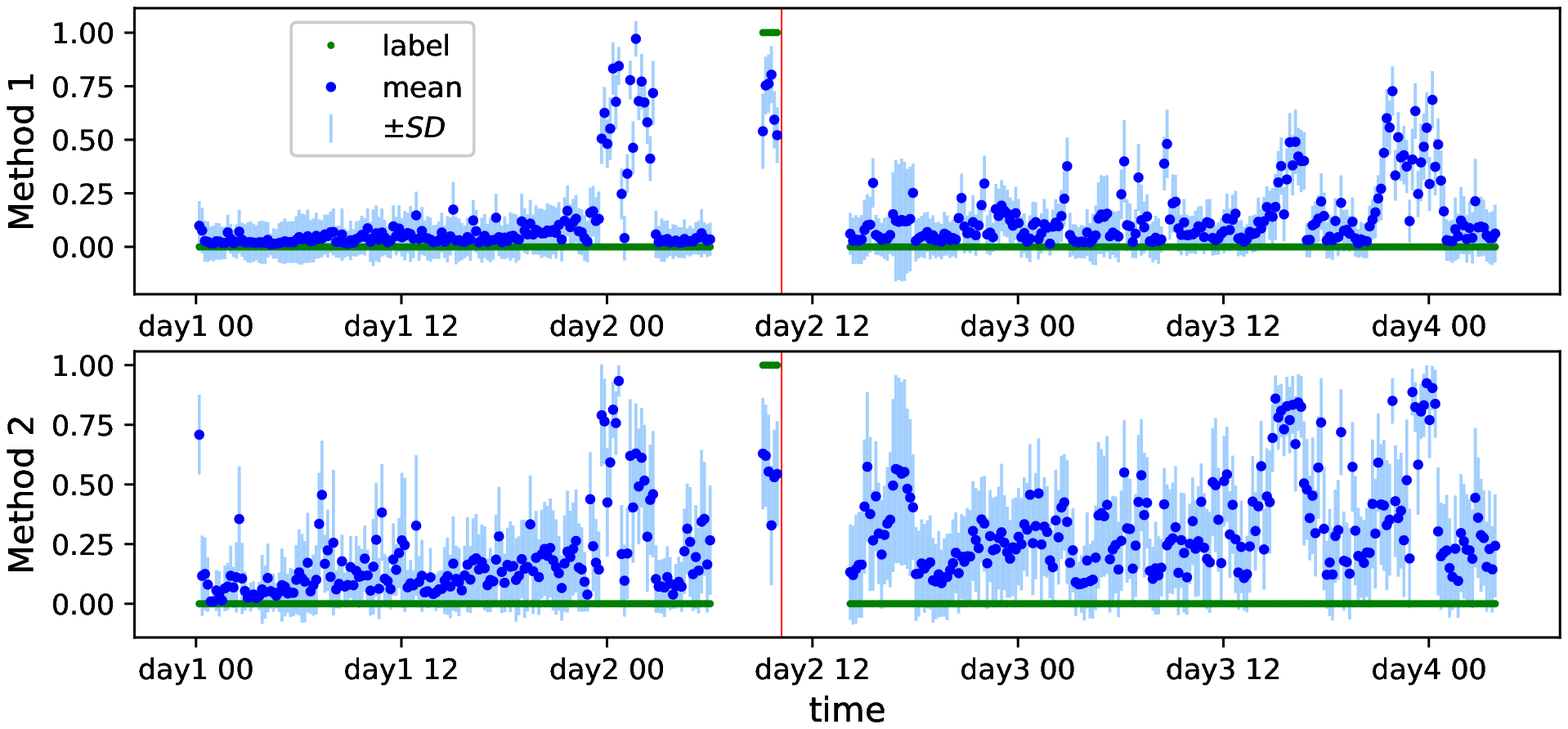}}
	\caption{Predictions for patients C-E of dataset 1. Seizure onset is marked as red line.}	
	\label{fig:ukdd-prb-part2}
\end{figure}

\newpage
\subsection*{Information Transfer Plots for all individuals of datasets 2 and 3}

\begin{figure}[H]
	\centering
    \subfigure[Dog 1]{\label{fig:inf-trans-dog1-supp}\includegraphics[width=0.49\columnwidth]{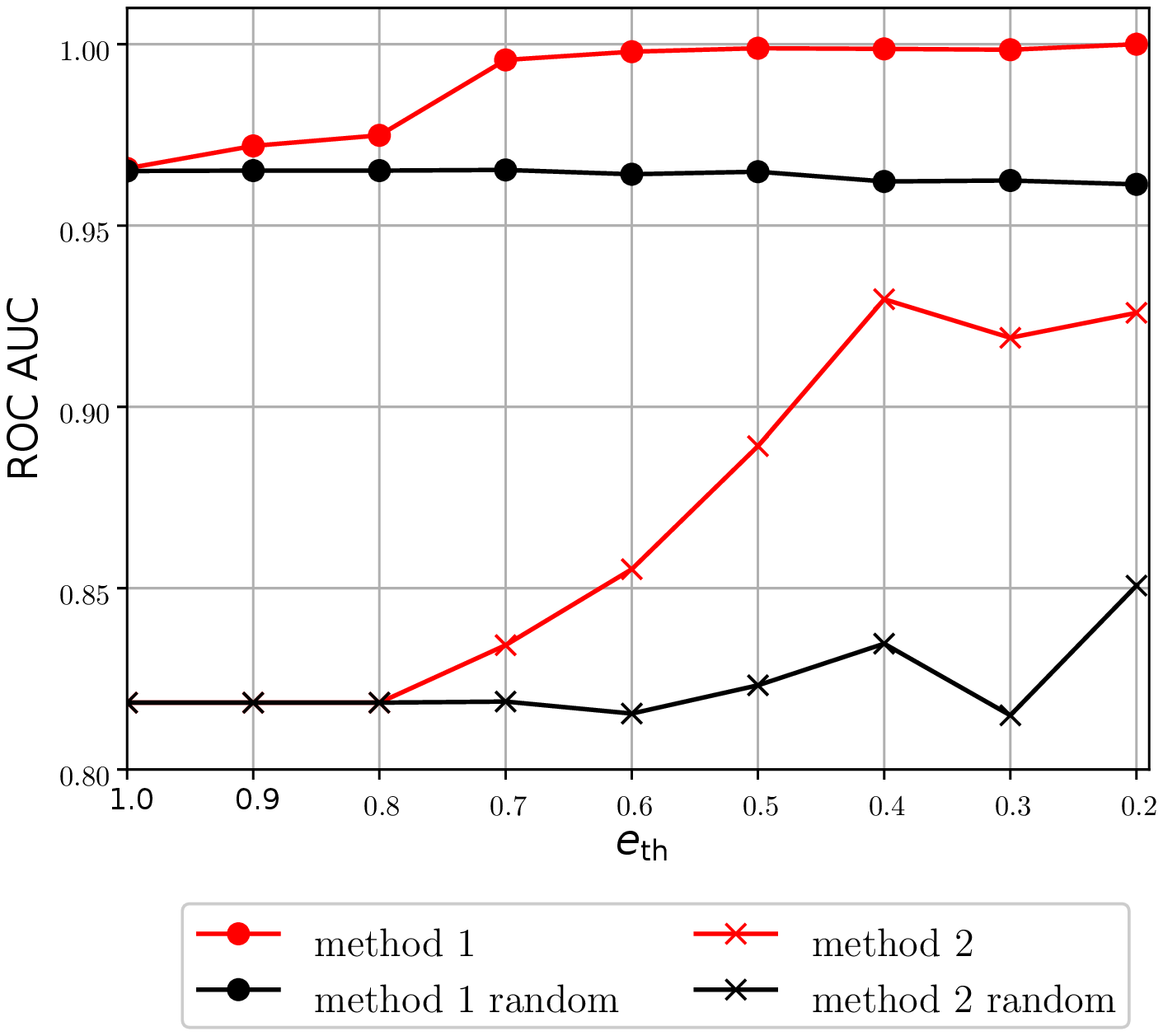}}\hfill
    \subfigure[Dog 2]{\label{fig:inf-trans-dog2-supp}\includegraphics[width=0.49\columnwidth]{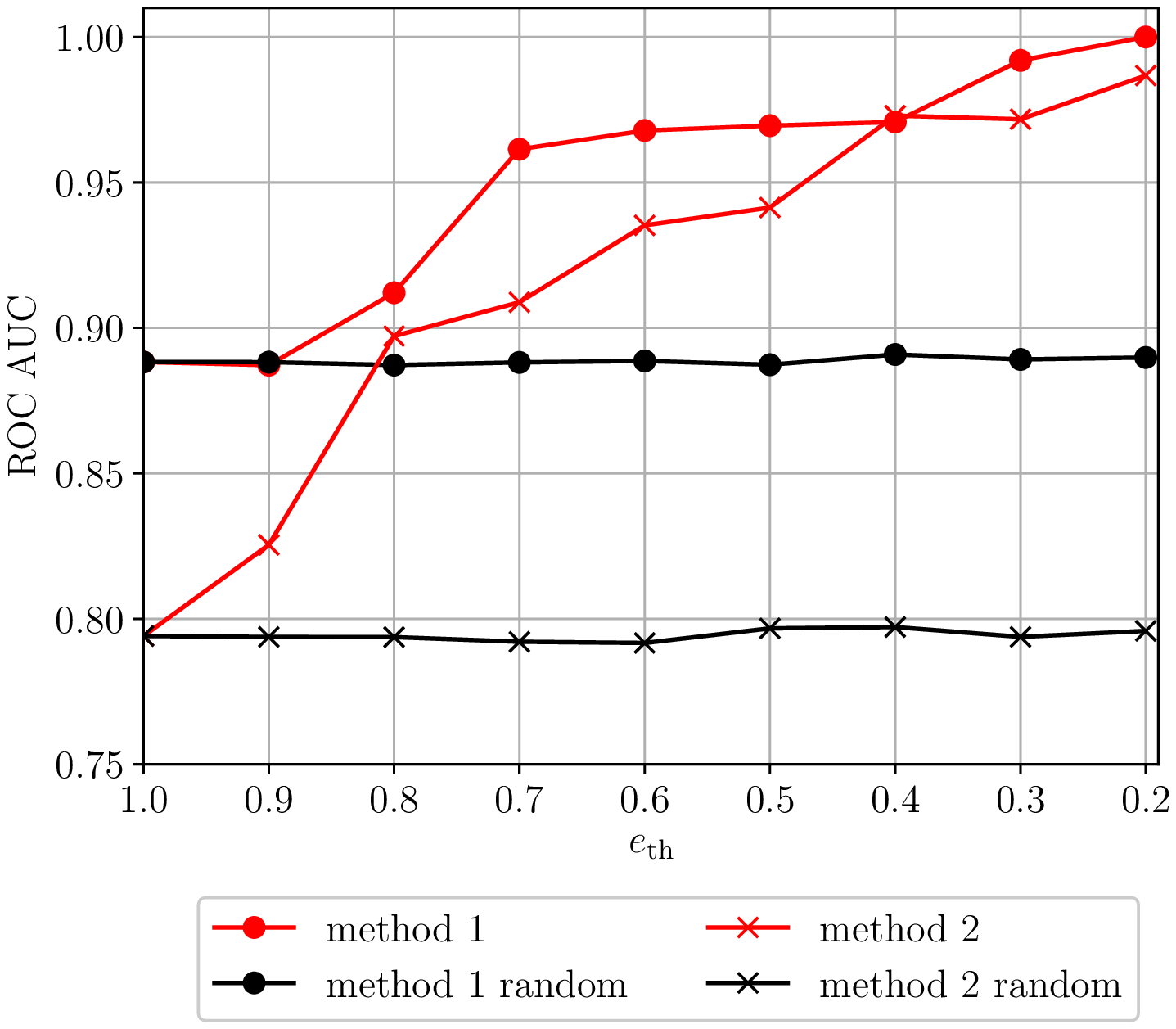}}
    \subfigure[Dog 3]{\label{fig:inf-trans-dog3-supp}\includegraphics[width=0.49\columnwidth]{data_inf-transf_dog3.eps}}\hfill
    \subfigure[Dog 4]{\label{fig:inf-trans-dog4-supp}\includegraphics[width=0.49\columnwidth]{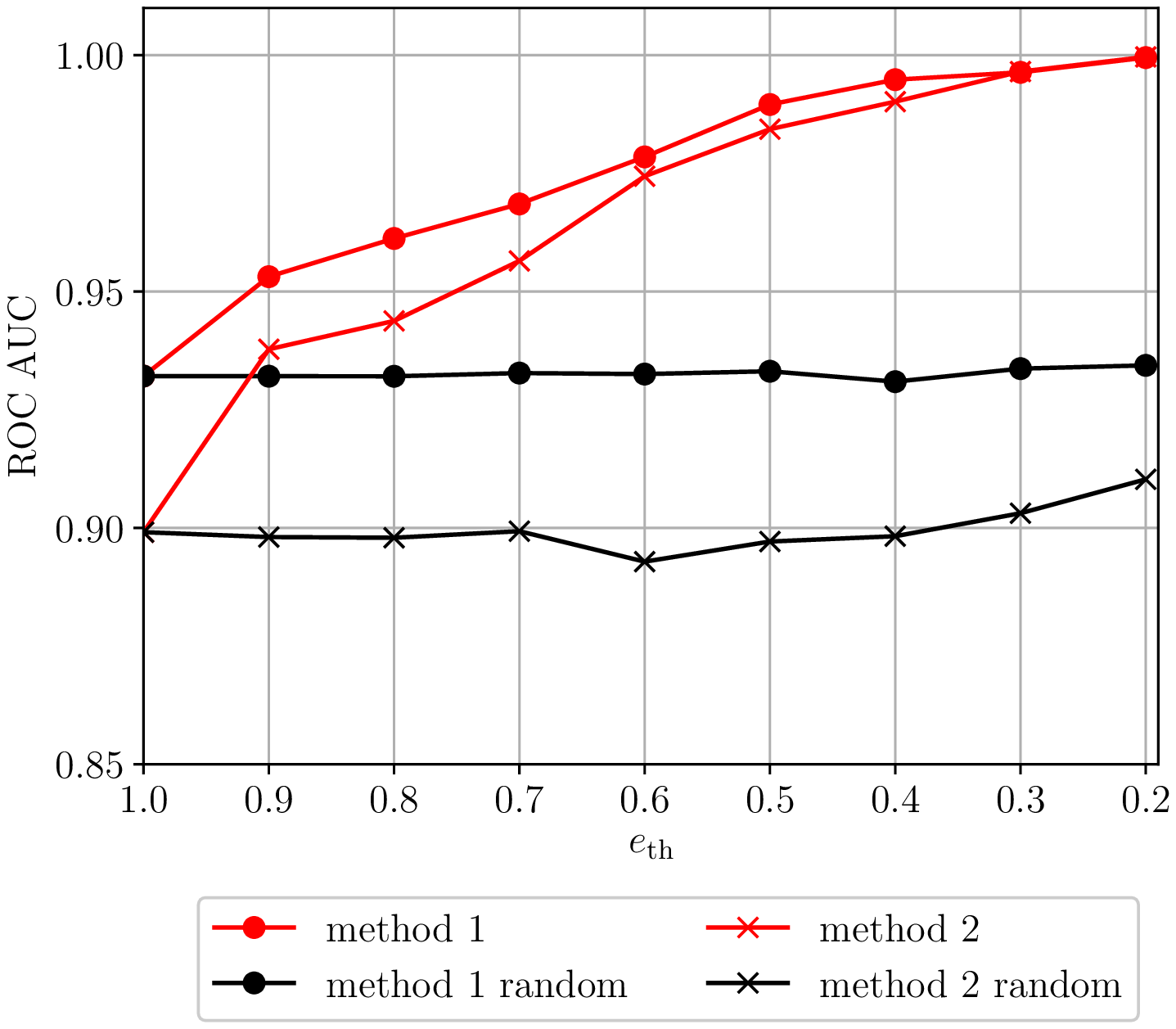}}
	\caption{Coherence of false predictions demonstrated by information transfer for all individuals of dataset 2.}
	\label{fig:inf-transf-dogs-supp}
\end{figure}

\newpage
\begin{figure}[H]
	\centering
    \subfigure[Patient 1]{\label{fig:inf-trans-pat1-supp}\includegraphics[width=0.49\columnwidth]{data_inf-transf_pat1.eps}}\hfill
    \subfigure[Patient 2]{\label{fig:inf-trans-pat2-supp}\includegraphics[width=0.49\columnwidth]{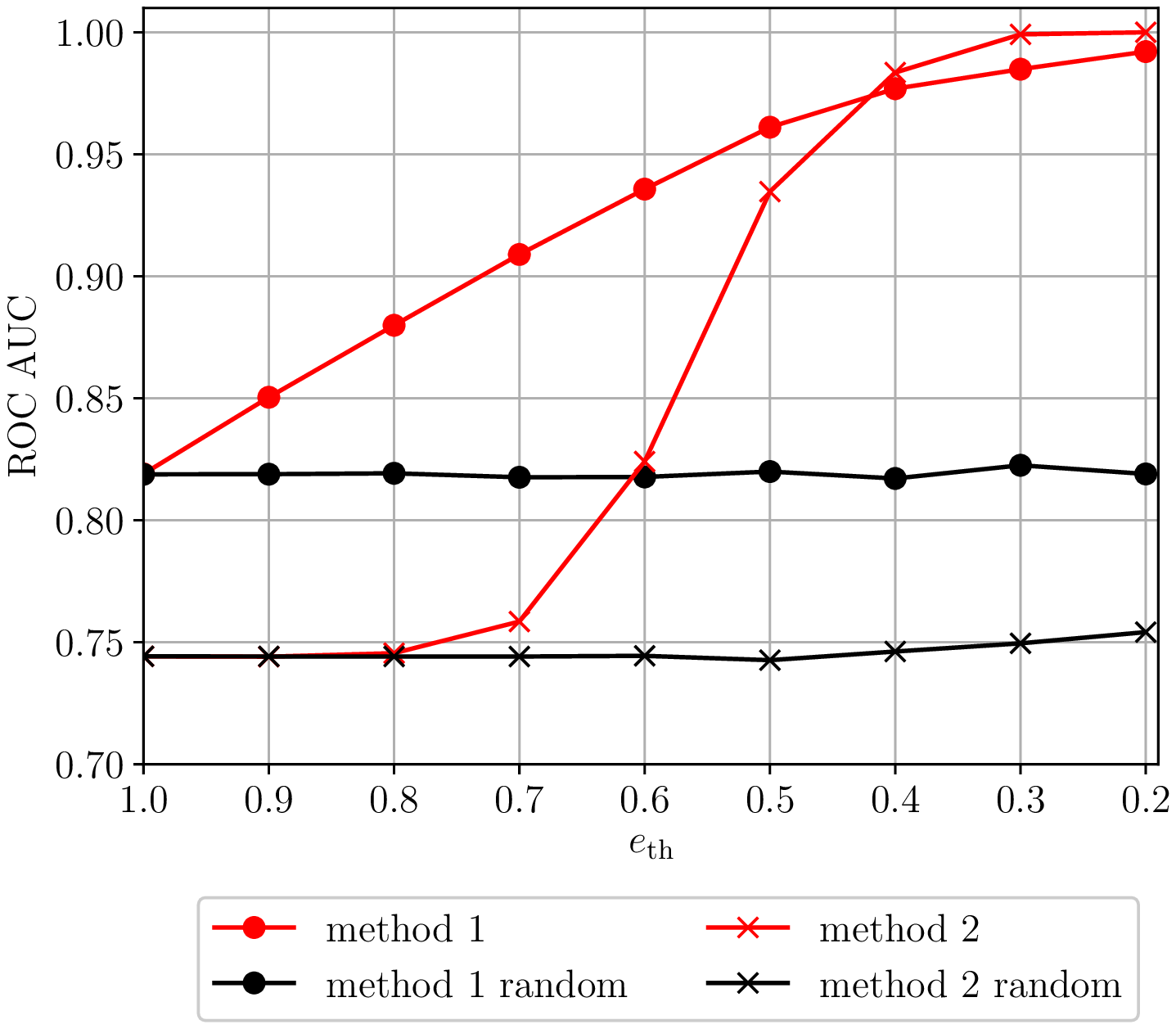}}
    \subfigure[Patient 3]{\label{fig:inf-trans-pat3-supp}\includegraphics[width=0.49\columnwidth]{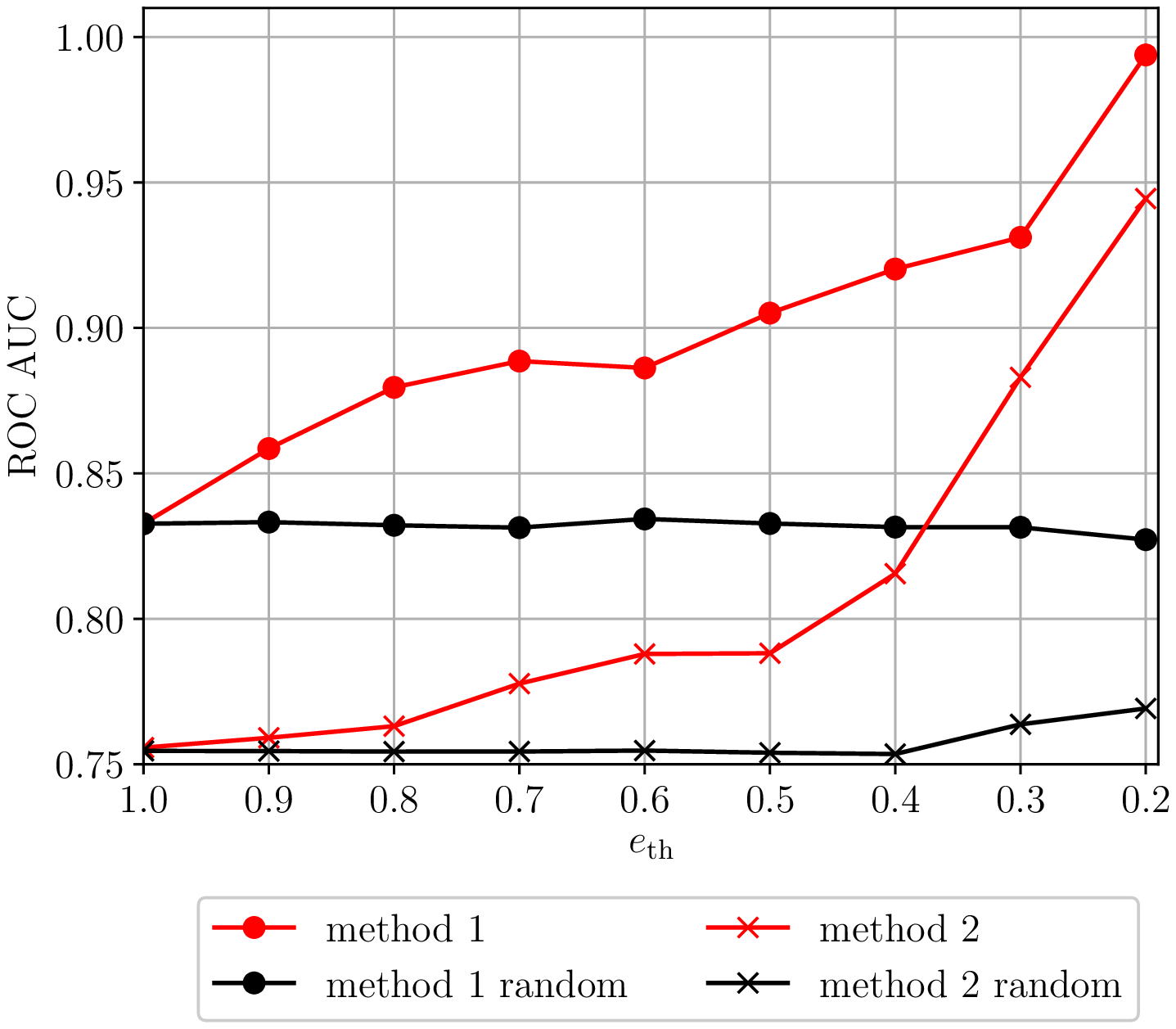}}
	\caption{Coherence of false predictions demonstrated by information transfer for all individuals of dataset 3.}
	\label{fig:inf-transf-pats-supp}
\end{figure}

\end{document}